%% file: arxiv.tex
\pdfoutput=1 % arXiv: force pdflatex (PDF/PNG figures)

\PassOptionsToPackage{table}{xcolor}

\documentclass[11pt,letterpaper,logo]{preprint}
\usepackage[utf8]{inputenc}
\usepackage[T1]{fontenc}
\usepackage{microtype}

\usepackage[numbers,sort]{natbib}

\usepackage{amsmath}
\usepackage{amssymb}
\usepackage{amsfonts}
\usepackage{nicefrac}

\usepackage{graphicx}
\usepackage{subcaption}
\usepackage{float}
\usepackage{wrapfig}
\usepackage{booktabs}
\usepackage{multirow}
\usepackage{makecell}

\usepackage{algorithm}
\usepackage{algpseudocode}

\usepackage{etoc}
\usepackage{url}
\usepackage{xurl}

\usepackage[disable]{todonotes}

\usepackage{tcolorbox}

\usepackage{hyperref}
\hypersetup{hidelinks}

\usepackage{doi}
\usepackage{cleveref}

\graphicspath{{./figures/}}

\expandafter\def\expandafter\UrlBreaks\expandafter{%
  \UrlBreaks\do\-%
}

\input{math_commands.tex}

\newcommand{\method}{\texttt{SCOUT}}

\definecolor{summarygray}{RGB}{245,245,245}
\definecolor{scoutsummary}{RGB}{235,242,250}
\definecolor{deepblue}{RGB}{18,68,150}

\definecolor{reinventpurple}{HTML}{8E44AD}
\definecolor{reinventdark}{HTML}{3F2A56}
\definecolor{reinventbg}{RGB}{248,244,252}

\newcommand{\res}[2]{%
  $#1{\scriptstyle\,\pm\,#2}$%
}

\newcommand{\bestres}[2]{%
  $\mathbf{#1}{\scriptstyle\,\pm\,#2}$%
}

\newcommand{\secondres}[2]{%
  $\underline{#1}{\scriptstyle\,\pm\,#2}$%
}

\makeatletter
\newcommand{\symfootnotetext}[2]{{%
  \renewcommand{\thefootnote}{#1}%
  \footnotetext{#2}%
}}
\makeatother

\colorlet{PaperTitleBack}{AWSViolet50}

\paperlogos{\includegraphics[height=22pt]{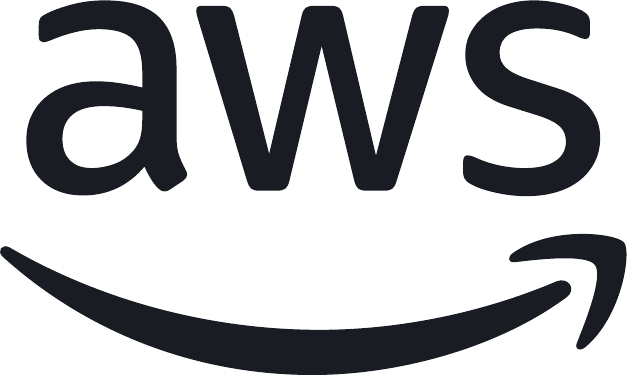}}

\title{On the Off-Policy Teacher in On-Policy Distillation}

\runningtitle{SCOUT: On the Off-Policy Teacher in On-Policy Distillation}

\metadata{Date}{\today}

\hypersetup{
  pdftitle={
    On the Off-Policy Teacher in On-Policy Distillation
  },
  pdfauthor={
    Langlin Huang,
    Hao Liu,
    Mononito Goswami,
    Xinyu Li,
    Prithwish Jana,
    Nikos Kanakaris,
    Patrick Blobaum,
    Purak Jain
  },
  pdfsubject={Machine Learning},
  pdfkeywords={
    on-policy distillation,
    knowledge distillation,
    reinforcement learning,
    language models
  }
}

\begin{document}

% ---------------------------------------------------------------------------
% Authors
% ---------------------------------------------------------------------------

\author{
  \normalfont
  \begin{minipage}{0.98\linewidth}
    \raggedright
    \setlength{\parskip}{0pt}

    \textbf{Langlin Huang}$^{1,\dagger,\ddagger}$,
    \textbf{Hao Liu}$^{2,\dagger}$,
    \textbf{Mononito Goswami}$^{2}$,
    \textbf{Xinyu Li}$^{3,\ddagger}$, \\
    \textbf{Prithwish Jana}$^{4,\ddagger}$,
    \textbf{Nikos Kanakaris}$^{2}$,
    \textbf{Patrick Bl\"obaum}$^{2}$,
    \textbf{Purak Jain}$^{2}$

    \par\vspace{7pt}

    $^{1}$Washington University in St. Louis
    \quad
    $^{2}$AWS AI Labs

    $^{3}$Carnegie Mellon University
    \quad
    $^{4}$Georgia Institute of Technology

  \end{minipage}
}

\begin{abstract}
\input{sections_arxiv/0Abstract}
\end{abstract}

\maketitle

% ---------------------------------------------------------------------------
% Author notes
% ---------------------------------------------------------------------------

{\let\thefootnote\relax
\footnotetext{%
  $^{\dagger}$Langlin Huang and Hao Liu contributed equally.
}}

{\let\thefootnote\relax
\footnotetext{%
  $^{\ddagger}$Langlin Huang, Xinyu Li, and Prithwish Jana were interns at
  AWS when this work was carried out.
}}

% Tag main-paper TOC entries separately so they can be hidden from the
% appendix-only table of contents.
\etocdepthtag.toc{mtmain}

% ===========================================================================
% Main paper
% ===========================================================================

\input{sections_arxiv/1Introduction}

\input{sections_arxiv/2Preliminary}

\input{sections_arxiv/3Method}

\input{sections_arxiv/4Experiment}

\input{sections_arxiv/5Analysis}

\input{sections_arxiv/5.5Relatedwork}

\input{sections_arxiv/6Conclusion}

% ===========================================================================
% References
% ===========================================================================

\clearpage

\bibliographystyle{unsrtnat}
\bibliography{arxiv}

\clearpage

% ===========================================================================
% Appendix
% ===========================================================================

\appendix

% Tag appendix entries separately from main-body entries.
\etocdepthtag.toc{mtappendix}

% Hide main-paper sections from the appendix TOC.
\etocsettagdepth{mtmain}{none}
\etocsettagdepth{mtappendix}{subsection}

% ---------------------------------------------------------------------------
% Appendix-only table of contents
% ---------------------------------------------------------------------------

% Appendix section entry: A, B, C, ...
\newcommand{\AppTocSection}{%
  \par
  \addvspace{0.9em}%
  \noindent
  \etoclink{%
    \makebox[2.8em][l]{\etocthenumber}%
    \textcolor{deepblue}{\textbf{\etocthename}}%
  }%
  \nobreak
  \leaders\hbox to 0.7em{\hss.\hss}\hfill
  \nobreak
  \makebox[2.2em][r]{\etocthepage}%
  \par
  \vspace{0.35em}%
}

% Appendix subsection entry: A.1, A.2, ...
\newcommand{\AppTocSubsection}{%
  \noindent
  \hspace*{1.8em}%
  \etoclink{%
    \makebox[3.4em][l]{\etocthenumber}%
    \textcolor{deepblue}{\etocthename}%
  }%
  \nobreak
  \leaders\hbox to 0.7em{\hss.\hss}\hfill
  \nobreak
  \makebox[2.2em][r]{\etocthepage}%
  \par
  \vspace{0.38em}%
}

\etocsetstyle{section}
  {}{}
  {\AppTocSection}
  {}

\etocsetstyle{subsection}
  {}{}
  {\AppTocSubsection}
  {}

\etoctoclines

\etocsettocstyle
  {%
    \section*{Appendix Contents}%
    \vspace{0.4em}
    \fontsize{11pt}{14pt}\selectfont
    \parskip=0pt
    \parfillskip=0pt
    \leftskip=1em
    \rightskip=1em
  }
  {%
    \vspace{0.8em}
  }

\tableofcontents

\clearpage

% ===========================================================================
% Appendix sections
% ===========================================================================

\input{sections_arxiv/7Appendix}

\end{document}

%% file: math_commands.tex
\usepackage{amsmath,amsfonts,bm}

\def\eqref#1{equation~\ref{#1}}
\def\1{\bm{1}}

\DeclareMathAlphabet{\mathsfit}{\encodingdefault}{\sfdefault}{m}{sl}
\SetMathAlphabet{\mathsfit}{bold}{\encodingdefault}{\sfdefault}{bx}{n}

%% file: sections_arxiv/0Abstract.tex
On-policy distillation (OPD) has recently emerged as a promising post-training paradigm in which the student learns from trajectories generated by its own policy under dense teacher supervision. However, OPD introduces a fundamental asymmetry: although the sampled trajectories are on-policy for the student, they are off-policy for the teacher. The teacher is typically optimized to continue from prefixes generated by its own policy, but during OPD it must instead supervise prefixes generated by the student. Empirically, we find that its continuation performance degrades as these prefixes grow longer. To address this issue, we propose \textbf{S}tudent-\textbf{CO}nditioned \textbf{U}pdates of the \textbf{T}eacher (\method), a co-training framework that adapts the teacher to student-generated prefixes. Alongside standard OPD updates, \method~periodically optimizes the teacher's conditional ability using reinforcement learning with verifiable rewards, where the teacher generates continuations from student prefixes and learns from outcome rewards. Controlled experiments show that \method~improves the teacher's ability to continue from student-generated prefixes, supporting the intended mechanism of student-conditioned teacher adaptation. Across multiple teacher--student configurations, model scales, and reasoning domains, \method~also consistently improves the effectiveness of on-policy distillation.

%% file: sections_arxiv/1Introduction.tex
\section{Introduction}
On-policy distillation (OPD) has emerged as a new method for Large Language Model post-training.
The student generates its own trajectories and a stronger teacher provides dense token-level supervision along the student-generated states \citep{agarwal2024onpolicydistillationlanguagemodels, gu2026minillmonpolicydistillationlarge}.
While on-policy trajectories benefit student training, OPD introduces a teacher-side distribution shift, where the teacher supervises prefixes sampled from the student-induced trajectory distribution rather than its own. As autoregressive generation proceeds, differences in intermediate decisions, reasoning patterns and errors, can accumulate, causing student-generated prefixes to become increasingly unlikely under the teacher policy. The teacher is therefore asked to provide next-token supervision in settings that it would rarely encounter during its own generation. 
Eventually, the teacher's token-level guidance may become less reliable given the accumulated mismatch, which in turn is harmful to OPD training~\citep{liu2026teachercanthelphere}.
Similar studies also suggest that the effectiveness of OPD depends not only on the teacher's standalone capability, but also on teacher-student compatibility and on the states at which supervision is provided \citep{li2026rethinkingonpolicydistillationlarge, zhou2026less, xing2026trust}.

Recent work mitigates this issue by regulating the supervision from a fixed teacher. Prefix-based methods restrict distillation to earlier portions of student trajectories, either to reduce the high cost of full rollouts or to avoid later prefixes where teacher supervision may deteriorate \citep{zhang2026fast, zhou2026less}. Other approaches retain a larger portion of the trajectory but modulate the teacher signal according to its uncertainty or teacher-student discrepancy, for example by changing the divergence objective, masking outlier tokens, or restricting updates to regions where the supervision is considered reliable \citep{jin2026entropyopd, xing2026trust}. Despite these differences, they share a common design choice: the teacher remains fixed, while the student update is adapted to selectively use its supervision. This raises a complementary question that has received less attention: rather than deciding when to trust a fixed teacher, \textbf{can we train the teacher itself to provide better supervision on student-generated prefixes?}

In this work, we make the teacher itself adaptive to the student-induced distribution. We introduce \textbf{\texttt{SCOUT}} (\textbf{S}tudent-\textbf{CO}nditioned \textbf{U}pdates of the \textbf{T}eacher), which explicitly trains the teacher on student-generated prefixes, turning teacher-side distribution shift from a fixed property of OPD into an optimization target. The student first samples a complete trajectory, from which a prefix is used to condition the teacher. The teacher generates a continuation from this prefix and is optimized using an outcome reward. The student, in turn, is trained with dense OPD over its full trajectory using the evolving teacher as supervision. This creates two coupled learning processes: the student learns from dense feedback on its own trajectories, while the teacher learns to better handle prefixes produced by the student. As the teacher improves on student-generated prefixes, it provides stronger supervision for subsequent student updates.

We evaluate \method~on mathematical reasoning across three teacher-student pairs, spanning different model scales, teacher training configurations, model families and additionally test its generalization to code generation. Across the three math settings, \method~consistently outperforms standard
OPD with a frozen teacher, improving average accuracy by $1.2$--$2.6$ points. On code generation, \method~improves the average score from $56.6$ to $59.7$, providing evidence that the approach extends beyond mathematical reasoning. Our analyses characterize where these gains come from. During training, the teacher progressively improves its ability to recover from student-generated prefixes. We further test whether the final gains could simply come from additional teacher training. A control with the same teacher updates but without student-prefix conditioning does not match SCOUT, showing that adaptation to student-generated states is central to the improvement. Finally, we also show that sparse teacher updates are sufficient to realize the benefit of student-conditioned adaptation and \method~is complementary to other OPD approaches.

\begin{tcolorbox}[
    enhanced,
    colback=AWSViolet50,
    colframe=PaperAccent,
    boxrule=0pt,
    leftrule=3pt,
    arc=6pt,
    left=7pt,
    right=7pt,
    top=6pt,
    bottom=6pt,
    before skip=8pt,
    after skip=8pt
]

{\papersans\color{PaperInk} Our contributions are threefold:}
\begin{itemize}
    \item We identify the teacher-side distribution shift as an important challenge in OPD and introduce teacher adaptation as a complementary optimization axis to regulating teacher supervision.
    \item We introduce \method, a simple and intuitive way to optimize along this new axis by training the teacher on student-generated prefixes with outcome-based RL, while retaining standard OPD for the student.
    \item We introduce a rigorous evaluation framework for \method, spanning two tasks and multiple teacher–student pairs, and use it to demonstrate that teacher adaptation generalizes across model scales, model families, and task domains.
\end{itemize}
% {\papersans\color{PaperInk} Contributions.}
% Our contributions are threefold: 
% (1) we identify the teacher-side distribution shift as an important challenge in OPD and introduce teacher adaptation as a complementary optimization axis to regulating teacher supervision; (2) we introduce \method, a simple and intuitive way to optimize along this new axis by training the teacher on student-generated prefixes with outcome-based RL, while retaining standard OPD for the student; (3) we introduce a rigorous evaluation framework for \method, spanning two tasks and multiple teacher–student pairs, and use it to demonstrate that teacher adaptation generalizes across model scales, model families, and task domains.

\end{tcolorbox}

%% file: sections_arxiv/2Preliminary.tex
\section{Preliminaries}
\label{sec:preliminaries}

\paragraph{On-Policy Distillation.}
Let $x \sim \mathcal{D}$ denote an input prompt sampled from a training distribution $\mathcal{D}$. Additionally, let $\pi_{\theta}$ and $\pi_{\phi}$ denote the student and teacher models, respectively. Knowledge distillation aims to improve the student $\pi_{\theta}$ by transferring knowledge from the teacher $\pi_{\phi}$. Given a response $y=(y_1,\ldots,y_T)$, the student and teacher define next-token distributions $\pi_{\theta}(\cdot \mid x,y_{<t})$ and $\pi_{\phi}(\cdot \mid x,y_{<t})$, where $y_{<t}=(y_1,\ldots,y_{t-1})$ denotes the response prefix preceding token $y_t$.

In off-policy distillation, training responses are typically generated by the teacher or drawn from a fixed dataset. Consequently, the student is trained on prefixes whose distribution may differ from the prefixes induced by its own autoregressive generation at inference time. This distribution mismatch is a form of exposure bias~\citep{ranzato2016sequenceleveltrainingrecurrent} and can limit the effectiveness of distillation.

On-policy distillation addresses this mismatch by collecting responses from the current student policy $y^S \sim \pi_{\theta}(\cdot \mid x)$. At each student-generated prefix $y^S_{<t}$, the teacher provides a next-token distribution $\pi_{\phi}(\cdot \mid x,y^S_{<t})$, yielding dense token-level supervision on states actually visited by the student. The standard OPD objective is
\begin{equation*}
\mathcal{L}_{\mathrm{OPD}}(\theta;\phi)
=
\mathbb{E}_{
x \sim \mathcal{D},
y^S \sim \pi_{\theta}(\cdot \mid x)
}
\left[
\frac{1}{|y^S|}
\sum_{t=1}^{|y^S|}
D\left(
\pi_{\theta}(\cdot \mid x,y^S_{<t}),
\pi_{\phi}(\cdot \mid x,y^S_{<t})
\right)
\right],
\end{equation*}
where $D(\cdot,\cdot)$ denotes a measure of divergence between the student and teacher next-token distributions, such as forward or reverse Kullback--Leibler (KL) divergence or Jensen--Shannon divergence. Unlike off-policy distillation, OPD trains the student on prefixes induced by its current policy. It therefore reduces the mismatch between training and inference-time contexts while retaining the teacher's dense distributional supervision.

\paragraph{Group Relative Policy Optimization.}
\label{sec:grpo-preliminaries}
Like OPD, GRPO~\citep{shao2024deepseekmath} trains on responses sampled from the current policy, but uses sequence-level rewards instead of teacher token distributions. Given a context $c$, GRPO samples $G$ responses from a policy and assigns each a reward $r_i=R(c,z_i)$, where $z_i \sim \pi_{\theta_{\mathrm{old}}}(\cdot \mid c),
\quad i=1,\ldots,G$. 
These rewards are normalized within the group to obtain a relative advantage $
\widehat{A}_i
=
\frac{r_i-\operatorname{mean}(\textbf{r})}
{\operatorname{std}(\textbf{r})}.
$
For each token, GRPO computes the importance ratio between the current and behavior policies,
$
\rho_{i,t}(\theta)
=
\frac{
\pi_{\theta}(z_{i,t}\mid c,z_{i,<t})
}{
\pi_{\theta_{\mathrm{old}}}(z_{i,t}\mid c,z_{i,<t})
},
$
and optimizes a clipped policy-gradient objective with optional KL regularization to a reference policy.

\[\mathcal{J}_{\mathrm{GRPO}}(\theta)
=
\mathbb{E}
\left[
\frac{1}{G}
\sum_{i=1}^{G}
\frac{1}{|z_i|}
\sum_{t=1}^{|z_i|}
\min\left(
\rho_{i,t}(\theta)\widehat{A}_i,
\operatorname{clip}
(\rho_{i,t}(\theta),1-\varepsilon,1+\varepsilon)
\widehat{A}_i
\right)
-
\beta D_{\mathrm{KL}}
\left(
\pi_{\theta}\,\|\,\pi_{\mathrm{ref}}
\right)
\right].\]

%% file: sections_arxiv/3Method.tex
\section{SCOUT: Student-COnditioned Updates of the Teacher}
\label{sec:method}
 \subsection{The Off-Policy Teacher in OPD}
\label{subsec:offpolicy-teacher}
OPD enables on-policy student training by optimizing the student on its own generated trajectories, while the teacher remains off-policy, providing supervision on trajectories it did not generate. Specifically, the teacher is primarily optimized to model $\pi_\phi(\cdot \mid x, y^T_{<t})$, where the preceding context follows its own trajectory distribution. However, during OPD, it must provide supervision under $\pi_\phi(\cdot \mid x, y^S_{<t})$, conditioned on prefixes generated by the student. This discrepancy exposes the teacher to contexts that may be unlikely under its native generation distribution. As differences in intermediate reasoning accumulate, this mismatch may become more pronounced later in the trajectory. We therefore hypothesize that the teacher becomes less effective at providing supervision on longer student-generated prefixes. We refer to this phenomenon as the \textbf{off-policy teacher} issue.

\begin{wrapfigure}[22]{r}{0.48\textwidth}
% \vspace{-1em}
  \centering
  \includegraphics[width=0.45\textwidth]{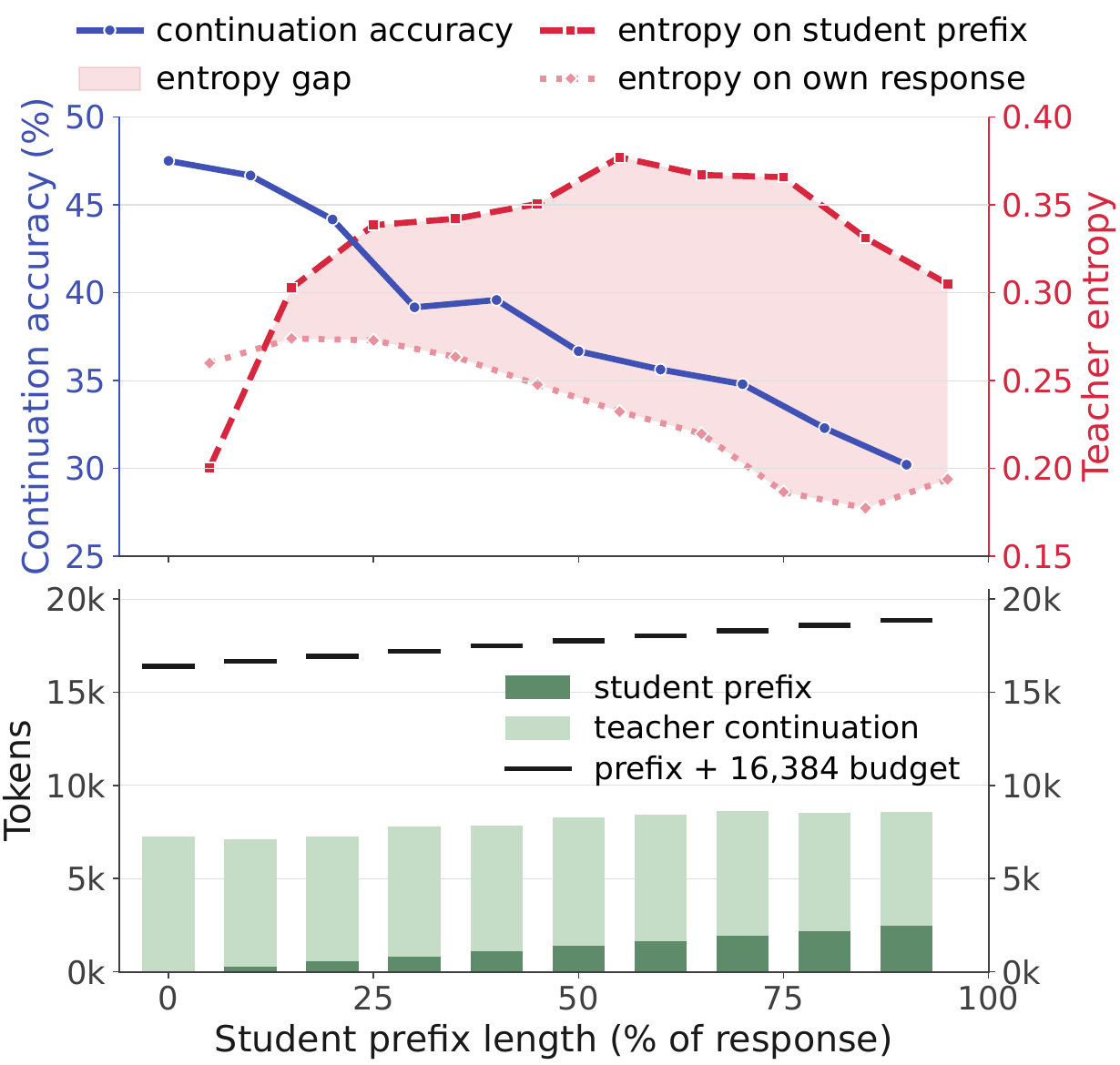}
  \caption{Teacher continuation from student-generated prefixes. As the student prefix grows longer, final-answer accuracy declines and a persistent entropy gap emerges between student and teacher prefixes, indicating increasing teacher uncertainty under student-generated contexts.}
  \label{fig:preliminary}
\end{wrapfigure}

To empirically examine the teacher's behavior under student-generated contexts, we evaluate its ability to continue from student prefixes $\pi_\phi(\cdot \mid x, y^S_{<t})$. For each question in AIME 2025, the student generates 4 complete responses, which we truncate at prefix ratios ranging from $10\%$ to $90\%$. From each truncated prefix, we sample 4 independent teacher continuations and report their average final-answer accuracy. This measures how the teacher's continuation ability changes as it is conditioned on increasingly long student-generated prefixes. We use Qwen3-1.7B in non-thinking mode as the student and
Qwen3-4B-Instruct-2507 as the teacher, with a maximum response length of
16,384 tokens for both, giving the teacher sufficient generation budget to recover from potentially suboptimal student prefixes.

As shown by the blue line in Figure~\ref{fig:preliminary}, teacher continuation accuracy generally decreases as the student prefix becomes longer. Importantly, the token-length statistics shown by the green bars indicate that the generated continuations remain well within the available response budget, ruling out insufficient generation length as the cause of this degradation. The decline therefore reflects a genuine deterioration in the teacher's ability to continue from $y^S_{<t}$. 
% These results support the hypothesis that longer student prefixes increasingly move the teacher away from the contexts encountered under its own generation distribution, making it a less reliable source of supervision at later positions along student-generated trajectories.

We further compare the teacher's average next-token entropy when conditioned on its own prefixes, $\pi_\phi(\cdot \mid x, y^T_{<t})$, versus student prefixes, $\pi_\phi(\cdot \mid x, y^S_{<t})$, as shown by the red dashed lines in Figure~\ref{fig:preliminary}. As the trajectory grows longer, entropy on teacher-generated prefixes decreases, whereas entropy on student-generated prefixes increases and then remains high, creating a persistent gap between the two. Together with the decline in continuation accuracy, this shows that the teacher has greater difficulty operating on longer student-generated contexts.

% As a result, the teacher produces an increasingly unreliable predictive distribution and is less able to provide the student with accurate and effective supervision, particularly at later positions along long student-generated trajectories.teacher along long student trajectories.

\subsection{Method}
\label{sec:method-overview}
% On longer student prefixes, $\pi_\phi$ exhibits (1) higher prediction agreement ratio with $\pi_\theta$, (2) lower next-token entropy, and (3) lower continuation accuracy. These indicates that the teacher policy tends to trust and follow student-generated prefixes instead of providing correction. 

\begin{figure}[t]
    \centering
    \includegraphics[width=\linewidth]{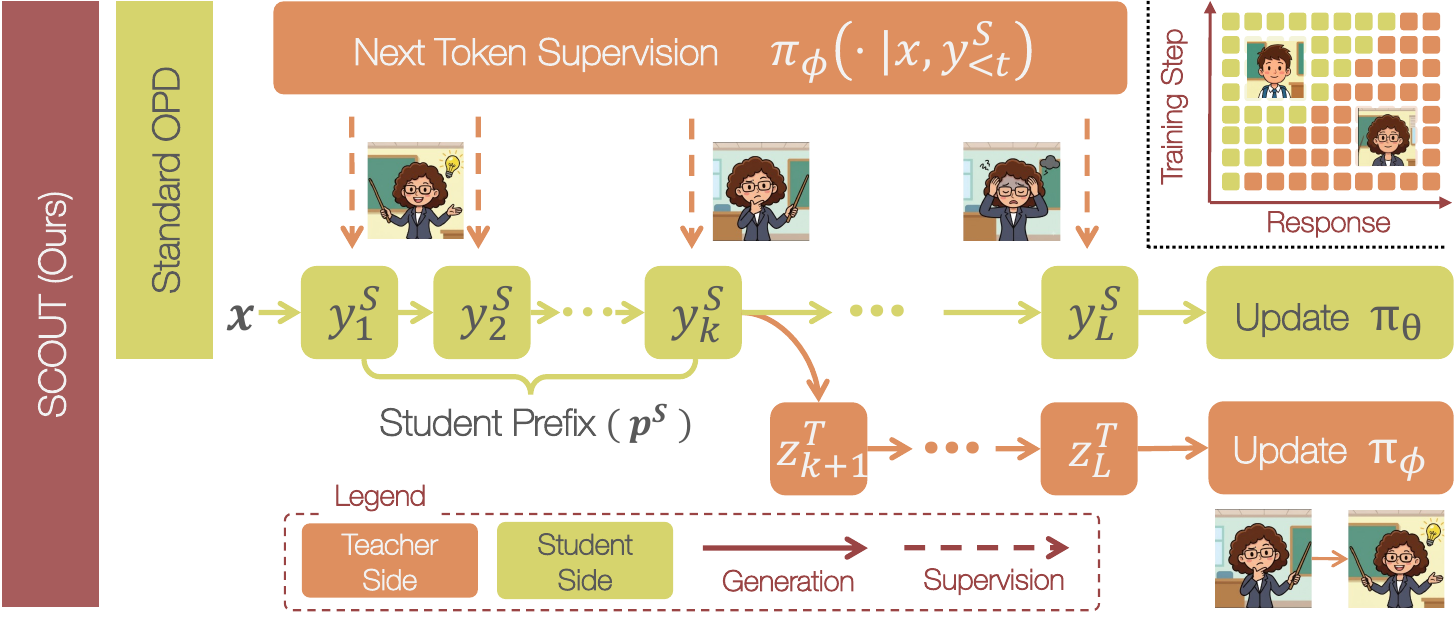}
    \caption{The overview of \method~and its comparison with standard OPD. While standard OPD suffers from unreliable teacher supervision at long prefixes, \method~periodically trains the teacher $\pi_\phi$ given the student prefix $p^S$. As shown in the figure, \method~is a combination of standard OPD and teacher-side RL training. $\pi_\phi$ is synchronized after its update to provide supervision for the next OPD training steps. Upper right subfigure: $\pi_\phi$ becomes gradually better at student prefixes along training. So we linearly increase the student prefix ratio according to training steps to increase the difficulty level.
    }
    \label{fig:method}
\end{figure}

\method~retains the student-side training of standard OPD and augments it with periodic teacher adaptation, as illustrated in Figure~\ref{fig:method}. In standard OPD, the student generates a trajectory $y^S=(y^S_1,\ldots,y^S_T)\sim\pi_\theta(\cdot\mid x)$, and the teacher provides dense next-token supervision along this student-generated trajectory. \method~leaves this OPD update unchanged and adds a teacher-side RL update conditioned on prefixes from the same student trajectories.
Specifically, we select a split point $k$ and define the student prefix $p^S=y^S_{1:k}$. Conditioned on the prompt and $p^S$, the teacher samples $K$ continuations,
\begin{equation}
z_i^T \sim \pi_\phi(\cdot\mid x,p^S), \qquad i=1,\ldots,K.
\end{equation}
Each response receives a verifiable outcome reward, and the teacher is updated with RL using it. Gradients are applied only to the teacher-generated continuation tokens. The updated teacher is then synchronized back to provide supervision for subsequent OPD updates.
This yields two coupled learning processes: the student learns from dense teacher supervision on its own trajectories, while the teacher learns from outcome feedback on continuations of student-generated prefixes.

Since the student policy $\pi_\theta$ evolves throughout training, the distribution of student-generated contexts evolves with it. The teacher must therefore be updated periodically to remain effective on the contexts visited by the current student. For computational efficiency, we update $\pi_\phi$ once every $f$ OPD steps, where $f$ represents the update interval. 
We further increase the student-prefix ratio linearly over training, as illustrated in the upper right subfigure of Figure~\ref{fig:method}. 
% This creates a simple curriculum for teacher adaptation: the teacher first learns from shorter prefixes that remain closer to its original distribution, then progressively adapts to longer and more challenging student-generated contexts.
As observed in Section~\ref{subsec:offpolicy-teacher}, longer student prefixes are more difficult for the teacher to continue from. We therefore begin teacher adaptation with shorter prefixes and gradually expose the teacher to longer and more challenging portions of the student trajectory as training progresses.

%% file: sections_arxiv/4Experiment.tex
\section{Experiments}
\label{sec:experiments}

\subsection{Experimental Setup}
\label{sec:experimental-setup}

We evaluate \method~on mathematical reasoning and code generation domains across three model pairs to test its generalization across task domains and model families.

\paragraph{Model configurations and training data.}
For the mathematical reasoning task, we experiment with (1) Qwen3-8B-DAPO as the teacher, obtained by training Qwen3-8B with GRPO on DAPO-MATH~\citep{yu2025dapoopensourcellmreinforcement} for three epochs\footnote{Because Qwen3-8B and Qwen3-1.7B are distilled from the same larger model~\citep{yang2025qwen3technicalreport}, we further train the 8B model to provide additional task-specific capability for distillation~\citep{li2026rethinkingonpolicydistillationlarge}.}, and Qwen3-1.7B as the student; (2) Qwen3-4B-Instruct-2507 as the teacher and Qwen3-1.7B as the student; and (3) Skywork-OR1-Math-7B~\citep{he2025skywork} as the teacher with DeepSeek-R1-Distill-Qwen-1.5B~\citep{deepseekai2025deepseekr1incentivizingreasoningcapability} as the student. We use OpenR1-Math-46K-8192~\citep{yan2025learningreasonoffpolicyguidance} as the training dataset.
For the code generation task, we use Qwen3-4B-Instruct-2507 as the teacher and Qwen3-1.7B as the student and the TACO-Verified-7.5K subset of the DeepCoder training data~\citep{deepcoder2025} as the training dataset.

\paragraph{Baselines.}
We compare \textbf{\method} against standard \textbf{OPD} with a frozen teacher, outcome-based RL using \textbf{GRPO}, and several recent OPD variants. \textbf{ESR}~\citep{zhou2026less} restricts distillation to an early portion of the student trajectory to avoid supervision on increasingly off-policy prefixes. \textbf{Prune-OPD}~\citep{yang2026prune} dynamically restricts supervision according to local teacher--student compatibility. \textbf{Relay-OPD}~\citep{xu2026relay} allows the teacher to intervene during rollout generation. 

ESR uses a truncation length of 4096 following~\citet{xu2026relay}\footnote{The original truncation length 100 yields worse performance than 4096, so we choose the stronger one.}.
\method~uses the same student-side distillation objective as standard OPD, while additionally updating the teacher through GRPO on continuations conditioned on student-generated prefixes.

\paragraph{Evaluation.}
We evaluate on six math reasoning benchmarks: AIME 2024~\citep{aime24}, AIME 2025~\citep{aime25}, AMC 2023, HMMT February 2025~\citep{balunovic2026matharenaevaluatingllmsuncontaminated}, OlympiadBench~\citep{he2024olympiadbenchchallengingbenchmarkpromoting}, and MATH-500~\citep{lightman2023letsverifystepstep}.
For code generation, we evaluate on LiveCodeBench v5~\citep{jain2024livecodebench}, HumanEval+~\citep{chen2021evaluating, liu2023codegeneratedchatgptreally}, and MBPP~\citep{austin2021program}. 

To rigorously evaluate our methods, we account for randomness in both training and evaluation through repeated experiments. Each method is trained independently with three random seeds, and each resulting model is evaluated 4--32 times per benchmark, following benchmark-specific evaluation best practices.  We report the aggregate mean accuracy and standard deviation ($\pm$) in the main tables.  \textbf{Bold} and \underline{underline} denote the best and second-best results among OPD-based methods, respectively. Appendix~\ref{app:eval-details} and Appendix~\ref{app:full-results} provides full evaluation details, complete per-run results, and significance tests against the standard OPD baseline.

\subsection{Main Results}
\label{sec:main-results}

\paragraph{\method~scales across teacher sizes.}
We first test whether the gains from teacher adaptation persist as the teacher scales. Table~\ref{tab:qwen3-4b-main-results} (left) and Table~\ref{tab:qwen3-8b-math-results} use Qwen3-4B-Instruct-2507 and Qwen3-8B-DAPO teachers, respectively. \method~achieves mean accuracies of 51.4 and 51.6, improving over OPD by 2.2 and 2.6 points. It is also best or tied for best among OPD-based methods across all six benchmarks in both settings. These results show that \method~remains effective across teacher scales and training configurations.
\input{Tables/table_4b_main}

\input{Tables/table_8b_main}

\paragraph{\method~generalizes across task domains.}
Next, we evaluate the same Qwen3-4B-Instruct-2507 $\rightarrow$ Qwen3-1.7B pair on code generation. As shown in Table~\ref{tab:qwen3-4b-main-results}, \method~improves over OPD by 2.2 points on mathematical reasoning and 3.1 points on code generation. Relay-OPD, in contrast, remains competitive on math but drops to 12.1 mean accuracy on code, where its generations frequently contain syntax errors. This contrast suggests that modifying student rollouts can be sensitive to the task domain, while adapting the teacher to student-generated prefixes transfers more reliably.

\paragraph{\method~generalizes across model families.}
Our first two settings use Qwen3 teacher--student pairs. We therefore ask whether the same gains hold under a different model family, using Skywork-OR1-Math-7B to teach DeepSeek-R1-Distill-Qwen-1.5B. This setting is also more challenging for several baselines: GRPO collapses during training, while Prune-OPD substantially underperforms standard OPD. In contrast, \method~achieves a mean accuracy of 53.6, improving over OPD by 1.2 points and outperforming competing OPD methods overall. The gains therefore extend beyond the Qwen3 model family, even when the teacher--student pairing and training dynamics change.

Across these experiments, a consistent pattern emerges: adapting the teacher to student-generated prefixes improves OPD across changes in teacher scale, training configuration, model family, and task domain. \method~achieves the strongest aggregate performance among competing OPD methods in all three mathematical reasoning settings and in code generation, while several alternatives degrade sharply in specific settings. Together, these results suggest that student-conditioned teacher adaptation provides a simple and broadly effective way to improve on-policy distillation without relying on task- or model-specific interventions.

\input{Tables/table_skywork_7b_main}

%% file: Tables/table_4b_main.tex
\begin{table*}[t]
\centering
\caption{
Main results for Qwen3-4B-Instruct-2507 $\rightarrow$ Qwen3-1.7B.
\method~consistently outperforms competing OPD methods across mathematical reasoning and code generation, demonstrating that its gains generalize across domains.
}
\label{tab:qwen3-4b-main-results}

\setlength{\tabcolsep}{2.8pt}
\renewcommand{\arraystretch}{1.10}

\resizebox{\textwidth}{!}{%
\begin{tabular}{
    l
    *{7}{c}
    @{\hspace{7pt}}
    *{4}{c}
}
\toprule

\multirow{2}{*}{\textbf{Method}}
& \multicolumn{7}{c}{\textbf{Mathematical Reasoning}}
& \multicolumn{4}{c}{\textbf{Code Generation}} \\
\cmidrule(lr){2-8}
\cmidrule(lr){9-12}

& \makecell{AIME24\\{\small\textit{Avg@32}}}
& \makecell{AIME25\\{\small\textit{Avg@32}}}
& \makecell{AMC23\\{\small\textit{Avg@32}}}
& \makecell{HMMT\\{\small\textit{Avg@32}}}
& \makecell{Olym.\\{\small\textit{Avg@4}}}
& \makecell{MATH500\\{\small\textit{Avg@8}}}
& \textbf{Mean}
& \makecell{LCB v5\\{\small\textit{Avg@4}}}
& \makecell{HE+\\{\small\textit{Avg@16}}}
& \makecell{MBPP\\{\small\textit{Avg@8}}}
& \textbf{Mean} \\

\midrule

Qwen3-1.7B
& $13.3{\scriptstyle\pm4.2}$
& $10.3{\scriptstyle\pm3.2}$
& $46.8{\scriptstyle\pm4.9}$
& $5.4{\scriptstyle\pm2.6}$
& $42.3{\scriptstyle\pm1.5}$
& $73.1{\scriptstyle\pm1.5}$
& $31.9$
& $27.3{\scriptstyle\pm0.3}$
& $62.4{\scriptstyle\pm1.5}$
& $47.3{\scriptstyle\pm1.2}$
& $45.7$ \\

Qwen3-4B-Ins
& $54.9{\scriptstyle\pm4.6}$
& $44.5{\scriptstyle\pm4.8}$
& $92.4{\scriptstyle\pm3.0}$
& $28.0{\scriptstyle\pm4.5}$
& $72.0{\scriptstyle\pm1.6}$
& $93.5{\scriptstyle\pm0.7}$
& $64.2$
& $50.3{\scriptstyle\pm0.3}$
& $84.3{\scriptstyle\pm1.3}$
& $79.4{\scriptstyle\pm1.2}$
& $71.3$ \\

\midrule

GRPO
& $30.9{\scriptstyle\pm4.5}$
& $30.4{\scriptstyle\pm4.1}$
& $73.3{\scriptstyle\pm5.7}$
& $18.9{\scriptstyle\pm3.4}$
& $61.1{\scriptstyle\pm5.8}$
& $87.4{\scriptstyle\pm3.5}$
& $50.3{\scriptstyle\pm4.4}$
& $44.3{\scriptstyle\pm6.2}$
& $72.7{\scriptstyle\pm1.4}$
& $69.6{\scriptstyle\pm5.0}$
& $62.2{\scriptstyle\pm2.8}$ \\

\midrule

\rowcolor{summarygray}
OPD
& $33.8{\scriptstyle\pm3.0}$
& $24.3{\scriptstyle\pm1.8}$
& $\underline{73.9{\scriptstyle\pm1.9}}$
& $16.0{\scriptstyle\pm0.8}$
& $\underline{60.2{\scriptstyle\pm1.6}}$
& $\underline{87.0{\scriptstyle\pm1.1}}$
& $49.2{\scriptstyle\pm1.5}$
& $\underline{37.6{\scriptstyle\pm0.1}}$
& $\underline{69.7{\scriptstyle\pm0.7}}$
& $\underline{62.5{\scriptstyle\pm0.3}}$
& $\underline{56.6{\scriptstyle\pm0.3}}$ \\

ESR
& $31.9{\scriptstyle\pm0.9}$
& $25.5{\scriptstyle\pm1.1}$
& $70.6{\scriptstyle\pm2.6}$
& $16.0{\scriptstyle\pm1.4}$
& $59.0{\scriptstyle\pm0.5}$
& $86.3{\scriptstyle\pm0.6}$
& $48.2{\scriptstyle\pm0.5}$
& $36.0{\scriptstyle\pm2.3}$
& $68.1{\scriptstyle\pm1.0}$
& $61.3{\scriptstyle\pm2.2}$
& $55.1{\scriptstyle\pm1.7}$ \\

Prune-OPD
& $\underline{34.0{\scriptstyle\pm0.8}}$
& $\underline{27.3{\scriptstyle\pm0.8}}$
& $69.8{\scriptstyle\pm0.7}$
& $\underline{16.5{\scriptstyle\pm0.8}}$
& $58.7{\scriptstyle\pm0.8}$
& $85.9{\scriptstyle\pm0.2}$
& $48.7{\scriptstyle\pm0.2}$
& $36.8{\scriptstyle\pm0.2}$
& $69.4{\scriptstyle\pm0.7}$
& $61.2{\scriptstyle\pm0.4}$
& $55.8{\scriptstyle\pm0.4}$ \\

Relay-OPD
& $\mathbf{35.8{\scriptstyle\pm2.8}}$
& $24.8{\scriptstyle\pm1.2}$
& $\underline{73.9{\scriptstyle\pm1.2}}$
& $15.0{\scriptstyle\pm1.5}$
& $59.9{\scriptstyle\pm0.2}$
& $86.3{\scriptstyle\pm0.2}$
& $\underline{49.3{\scriptstyle\pm0.6}}$
& $2.3{\scriptstyle\pm0.4}$
& $14.3{\scriptstyle\pm1.6}$
& $19.6{\scriptstyle\pm0.9}$
& $12.1$ \\

\midrule

\rowcolor{scoutsummary}
\textbf{\method}
& $\mathbf{35.8{\scriptstyle\pm0.5}}$
& $\mathbf{28.8{\scriptstyle\pm0.6}}$
& $\mathbf{75.8{\scriptstyle\pm0.6}}$
& $\mathbf{17.2{\scriptstyle\pm1.9}}$
& $\mathbf{62.4{\scriptstyle\pm0.8}}$
& $\mathbf{88.4{\scriptstyle\pm0.3}}$
& $\mathbf{51.4{\scriptstyle\pm0.6}}$
& $\mathbf{40.6{\scriptstyle\pm2.5}}$
& $\mathbf{73.4{\scriptstyle\pm0.9}}$
& $\mathbf{65.2{\scriptstyle\pm2.6}}$
& $\mathbf{59.7{\scriptstyle\pm1.9}}$ \\

\bottomrule
\end{tabular}%
}

\end{table*}

% \symfootnotetext{$\dagger$}{
% GRPO collapses near the end of training, so the best checkpoint from each
% training run is used before averaging across the three independent runs.
% }

% \symfootnotetext{$\ddagger$}{
% Relay-OPD produces many syntax errors on code generation, resulting in poor performance.
% }

%% file: Tables/table_8b_main.tex
\begin{table*}[t]
\centering
\caption{
Mathematical reasoning with Qwen3-8B-DAPO $\rightarrow$ Qwen3-1.7B. \method~outperforms all other OPD-based methods across all six benchmarks, demonstrating that its gains generalize to larger model scales.
}
\label{tab:qwen3-8b-math-results}

\setlength{\tabcolsep}{2.8pt}
\renewcommand{\arraystretch}{1.10}

\resizebox{0.75\textwidth}{!}{%
\begin{tabular}{l*{7}{c}}
\toprule

\multirow{2}{*}{\textbf{Method}}
& \multicolumn{7}{c}{\textbf{Mathematical Reasoning}} \\
\cmidrule(lr){2-8}

& \makecell{AIME24\\{\small\textit{Avg@32}}}
& \makecell{AIME25\\{\small\textit{Avg@32}}}
& \makecell{AMC23\\{\small\textit{Avg@32}}}
& \makecell{HMMT\\{\small\textit{Avg@32}}}
& \makecell{Olym.\\{\small\textit{Avg@4}}}
& \makecell{MATH500\\{\small\textit{Avg@8}}}
& \textbf{Mean} \\

\midrule

Qwen3-1.7B
& $13.3{\scriptstyle\pm4.2}$
& $10.3{\scriptstyle\pm3.2}$
& $46.8{\scriptstyle\pm4.9}$
& $5.4{\scriptstyle\pm2.6}$
& $42.3{\scriptstyle\pm1.5}$
& $73.1{\scriptstyle\pm1.5}$
& $31.9$ \\

Qwen3-8B-DAPO
& $59.9{\scriptstyle\pm6.0}$
& $43.9{\scriptstyle\pm5.9}$
& $90.7{\scriptstyle\pm3.9}$
& $25.7{\scriptstyle\pm4.1}$
& $74.2{\scriptstyle\pm0.7}$
& $94.6{\scriptstyle\pm0.5}$
& $64.8$ \\

\midrule

GRPO
& $30.9{\scriptstyle\pm4.5}$
& $30.4{\scriptstyle\pm4.1}$
& $73.3{\scriptstyle\pm5.7}$
& $18.9{\scriptstyle\pm3.4}$
& $61.1{\scriptstyle\pm5.8}$
& $87.4{\scriptstyle\pm3.5}$
& $50.3{\scriptstyle\pm4.4}$ \\

\midrule

\rowcolor{summarygray}
OPD
& $33.4{\scriptstyle\pm0.8}$
& $\underline{27.8{\scriptstyle\pm1.0}}$
& $70.1{\scriptstyle\pm0.7}$
& $15.9{\scriptstyle\pm1.4}$
& $60.0{\scriptstyle\pm1.1}$
& $87.0{\scriptstyle\pm0.5}$
& $\underline{49.0{\scriptstyle\pm0.6}}$ \\

ESR
& $\underline{34.8{\scriptstyle\pm2.5}}$
& $26.7{\scriptstyle\pm1.9}$
& $69.5{\scriptstyle\pm2.0}$
& $15.8{\scriptstyle\pm0.2}$
& $\underline{60.6{\scriptstyle\pm0.4}}$
& $86.8{\scriptstyle\pm0.2}$
& $\underline{49.0{\scriptstyle\pm0.4}}$ \\

Prune-OPD
& $33.6{\scriptstyle\pm1.3}$
& $26.2{\scriptstyle\pm0.6}$
& $69.6{\scriptstyle\pm0.2}$
& $\underline{16.5{\scriptstyle\pm1.4}}$
& $58.9{\scriptstyle\pm0.6}$
& $86.4{\scriptstyle\pm0.4}$
& $48.5{\scriptstyle\pm0.2}$ \\

Relay-OPD
& $31.2{\scriptstyle\pm0.2}$
& $27.3{\scriptstyle\pm1.2}$
& $\underline{70.4{\scriptstyle\pm1.1}}$
& $15.4{\scriptstyle\pm1.1}$
& $59.5{\scriptstyle\pm0.7}$
& $\underline{87.2{\scriptstyle\pm0.3}}$
& $48.5{\scriptstyle\pm0.4}$ \\

\midrule

\rowcolor{scoutsummary}
\textbf{\method}
& $\mathbf{35.4{\scriptstyle\pm3.9}}$
& $\mathbf{31.2{\scriptstyle\pm0.7}}$
& $\mathbf{75.0{\scriptstyle\pm0.6}}$
& $\mathbf{17.7{\scriptstyle\pm1.2}}$
& $\mathbf{61.4{\scriptstyle\pm0.3}}$
& $\mathbf{89.0{\scriptstyle\pm0.7}}$
& $\mathbf{51.6{\scriptstyle\pm1.0}}$ \\

\bottomrule
\end{tabular}%
}
\vspace{-1em}
\end{table*}

% \symfootnotetext{$\dagger$}{
% GRPO collapses near the end of training, so the best checkpoint from each
% training run is used before averaging across the three independent runs.
% }

%% file: Tables/table_skywork_7b_main.tex
\begin{table*}[t]
\centering
\caption{Mathematical reasoning with Skywork-OR1-Math-7B $\rightarrow$ DeepSeek-R1-Distill-Qwen-1.5B. GRPO collapses and Prune-OPD degrades substantially in this setting, while \method~remains robust and outperforms baselines, demonstrating that its gains generalize across model families.
}
\label{tab:deepseek-math-results}

\setlength{\tabcolsep}{2.8pt}
\renewcommand{\arraystretch}{1.10}

\resizebox{0.85\textwidth}{!}{%
\begin{tabular}{l*{7}{c}}
\toprule

\multirow{2}{*}{\textbf{Method}}
& \multicolumn{7}{c}{\textbf{Mathematical Reasoning}} \\
\cmidrule(lr){2-8}

& \makecell{AIME24\\{\small\textit{Avg@32}}}
& \makecell{AIME25\\{\small\textit{Avg@32}}}
& \makecell{AMC23\\{\small\textit{Avg@32}}}
& \makecell{HMMT\\{\small\textit{Avg@32}}}
& \makecell{Olym.\\{\small\textit{Avg@4}}}
& \makecell{MATH500\\{\small\textit{Avg@8}}}
& \textbf{Mean} \\

\midrule

\makecell[l]{DeepSeek-R1-Distill-Qwen-1.5B}
& $28.1{\scriptstyle\pm4.7}$
& $23.8{\scriptstyle\pm4.7}$
& $70.6{\scriptstyle\pm4.8}$
& $13.9{\scriptstyle\pm4.8}$
& $52.9{\scriptstyle\pm1.4}$
& $83.2{\scriptstyle\pm0.8}$
& $45.4$ \\

\makecell[l]{Skywork-OR1-Math-7B}
& $57.7{\scriptstyle\pm5.4}$
& $46.4{\scriptstyle\pm4.8}$
& $90.2{\scriptstyle\pm2.7}$
& $27.3{\scriptstyle\pm3.9}$
& $73.4{\scriptstyle\pm0.9}$
& $94.8{\scriptstyle\pm0.7}$
& $65.0$ \\

\midrule

GRPO$^\dagger$
& $4.1{\scriptstyle\pm3.0}$
& $3.1{\scriptstyle\pm2.7}$
& $35.8{\scriptstyle\pm5.3}$
& $0.4{\scriptstyle\pm1.1}$
& $29.2{\scriptstyle\pm1.4}$
& $63.1{\scriptstyle\pm1.2}$
& $22.6$ \\

\midrule

\rowcolor{summarygray}
OPD
& $\underline{37.8{\scriptstyle\pm0.6}}$
& $\mathbf{30.0{\scriptstyle\pm0.5}}$
& $\underline{77.9{\scriptstyle\pm0.6}}$
& $\mathbf{19.1{\scriptstyle\pm0.1}}$
& $\underline{60.6{\scriptstyle\pm0.2}}$
& $\underline{89.1{\scriptstyle\pm0.1}}$
& $\underline{52.4{\scriptstyle\pm0.2}}$ \\

ESR
& $32.6{\scriptstyle\pm2.0}$
& $27.3{\scriptstyle\pm0.1}$
& $71.6{\scriptstyle\pm1.7}$
& $17.3{\scriptstyle\pm1.1}$
& $56.9{\scriptstyle\pm0.4}$
& $87.5{\scriptstyle\pm0.1}$
& $48.8{\scriptstyle\pm0.4}$ \\

Prune-OPD
& $21.6{\scriptstyle\pm1.0}$
& $18.8{\scriptstyle\pm0.4}$
& $56.8{\scriptstyle\pm0.3}$
& $11.2{\scriptstyle\pm0.3}$
& $40.5{\scriptstyle\pm0.7}$
& $72.0{\scriptstyle\pm0.4}$
& $36.8{\scriptstyle\pm0.1}$ \\

Relay-OPD
& $34.7{\scriptstyle\pm1.8}$
& $\underline{28.3{\scriptstyle\pm1.0}}$
& $74.6{\scriptstyle\pm1.6}$
& $16.4{\scriptstyle\pm0.8}$
& $58.9{\scriptstyle\pm1.0}$
& $88.3{\scriptstyle\pm0.2}$
& $50.2{\scriptstyle\pm0.6}$ \\

\midrule

\rowcolor{scoutsummary}
\textbf{\method}
& $\mathbf{39.8{\scriptstyle\pm0.7}}$
& $\mathbf{30.0{\scriptstyle\pm0.5}}$
& $\mathbf{81.1{\scriptstyle\pm0.9}}$
& $\underline{18.7{\scriptstyle\pm0.3}}$
& $\mathbf{61.8{\scriptstyle\pm0.3}}$
& $\mathbf{90.0{\scriptstyle\pm0.0}}$
& $\mathbf{53.6{\scriptstyle\pm0.2}}$ \\

\bottomrule
\end{tabular}%
}
\end{table*}

\symfootnotetext{$\dagger$}{GRPO collapses during training on DeepSeek-R1-Distill-Qwen-1.5B; similar instability has been reported with the same framework~\citep{verl2738}.}

%% file: sections_arxiv/5Analysis.tex
\section{Analysis}
\label{sec:analysis}

Next, we study how teacher adaptation changes the training dynamics. We focus on three questions: \textcolor{deepblue}{\textbf{(RQ1)}} Does teacher adaptation improve performance on student-generated prefixes? \textcolor{deepblue}{\textbf{(RQ2)}} Are the gains specific to adapting the teacher on student-generated prefixes? \textcolor{deepblue}{\textbf{(RQ3)}} How frequently should the teacher adapt to the evolving student? Finally, we evaluate under similar computational cost and whether \method~complements alternative approaches on improving teacher supervision.

\paragraph{\textcolor{deepblue}{(RQ1)} \method~adapts the teacher to student-generated prefixes.}
We study how the teacher changes on student-generated prefixes during training in the Qwen3-4B-Instruct-2507 $\rightarrow$ Qwen3-1.7B math setting on AIME 2025. We first examine continuation ability. The student generates a complete reasoning trajectory, from which we select prefixes ending at different points and ask the teacher to continue reasoning.

\begin{figure}[!h]%{r}{0.7\textwidth}
    % \vspace{-10pt}
    \centering
    \includegraphics[width=0.7\linewidth]{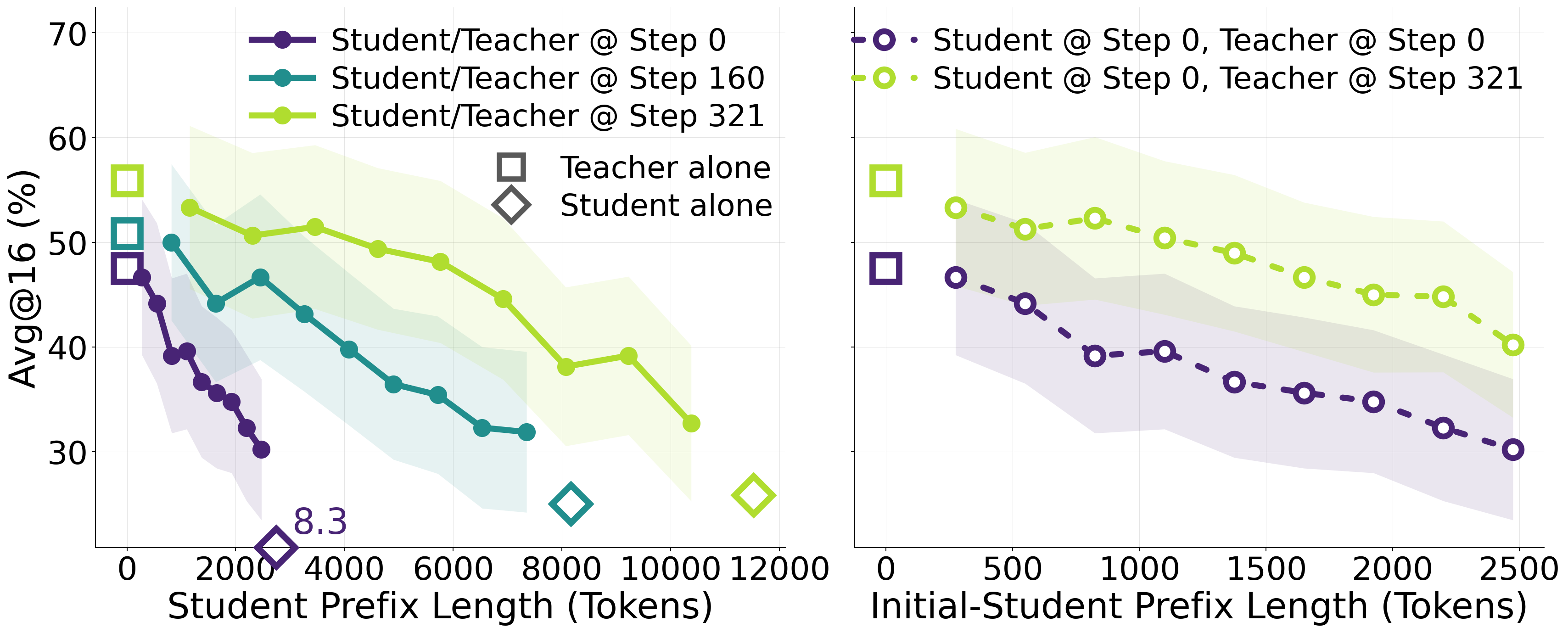}
    % \vspace{-20pt}
    \caption{\textit{\method~improves the teacher's ability to reason from student-generated prefixes.}
\textbf{Left:} Matched student--teacher checkpoints become more accurate across increasingly long prefixes as training progresses.
\textbf{Right:} Holding the student prefixes fixed, the trained teacher outperforms the initial teacher on the same inputs, showing that the gains reflect teacher adaptation rather than stronger student prefixes alone.}
    \label{fig:continuation}
    % \vspace{-16pt}
\end{figure}

Figure~\ref{fig:continuation} \textit{(left)} compares matched student and teacher checkpoints at initialization, midway through training (Step 160), and after one epoch (Step 321). As training progresses, the continuation curves shift upward: later checkpoint pairs achieve higher accuracy across increasingly long student-generated prefixes. This shows that the student--teacher pair becomes increasingly effective at reasoning from longer student-generated prefixes. However, since later teachers are evaluated on prefixes from later, stronger students, this comparison alone does not tell whether the improvement comes from teacher adaptation or simply from better student-generated prefixes. We therefore hold the student prefixes fixed. Figure~\ref{fig:continuation} \textit{(right)} compares the teacher at initialization and after one epoch of training on prefixes generated by the \textit{initial} student. The trained teacher achieves higher accuracy across all prefix lengths, showing that the improvement is not simply due to stronger student prefixes, but reflects improvement in the teacher itself.

\begin{wrapfigure}{r}{0.47\textwidth}
    \centering
    \includegraphics[width=\linewidth]{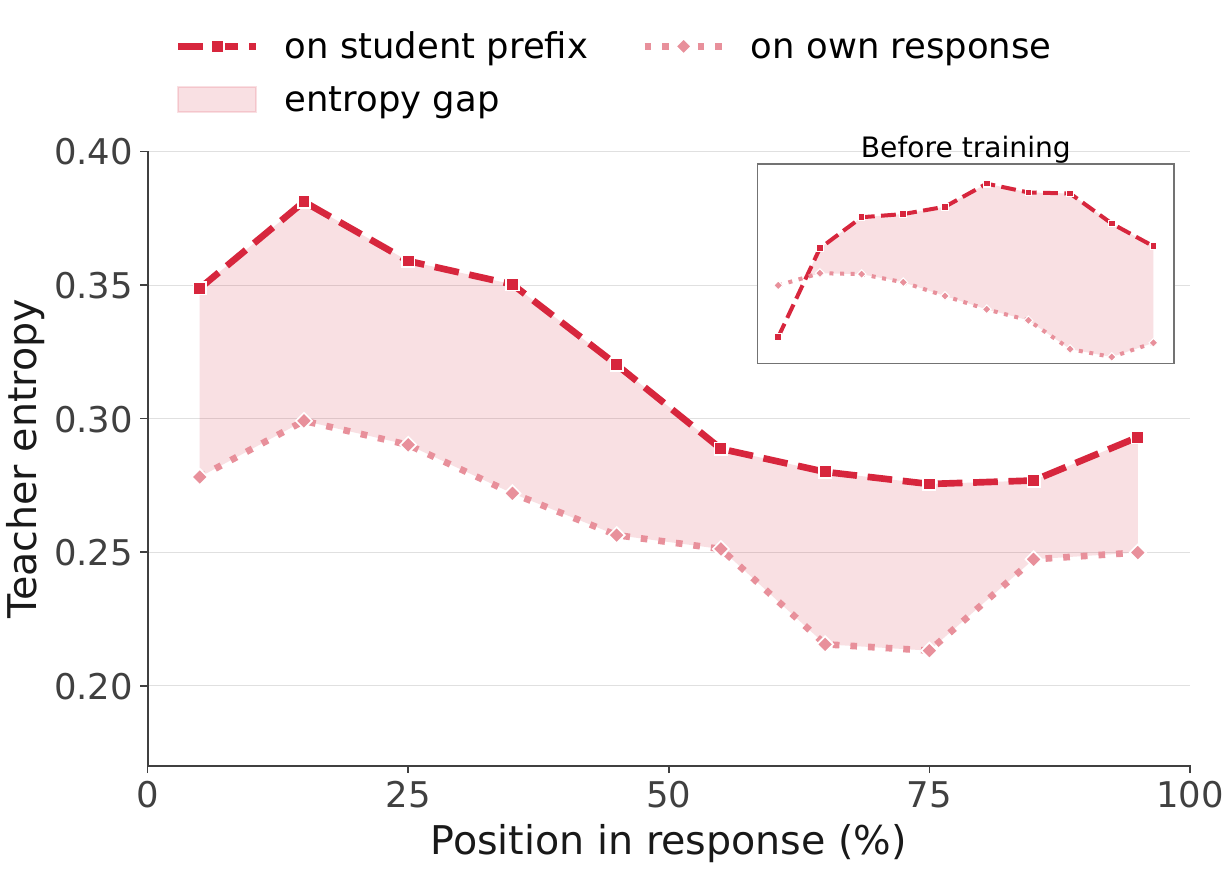}
    \caption{
    \textit{\method~reduces teacher uncertainty on student-generated prefixes.}
    After training, teacher entropy on student prefixes decreases over later response positions, narrowing the gap with entropy on the teacher's own responses. The inset shows the corresponding behavior before training.
    }
    \label{fig:entropy-after-training}
\end{wrapfigure}
We next examine whether \method~mitigates the teacher's uncertainty on student-generated contexts. Section~\ref{subsec:offpolicy-teacher} shows that before training, teacher entropy decreases along its own responses but remains high on student-generated prefixes. Figure~\ref{fig:entropy-after-training} revisits this diagnostic with the student--teacher pair after \method~training. Teacher entropy on student prefixes now decreases substantially over later response positions, and the gap between student-prefix and own-response entropy becomes smaller. This shows that \method~not only improves the teacher's continuation performance, but also makes it less uncertain on the student-generated contexts where it must provide supervision.

Together, these analyses show that \method~effectively adapts the teacher to the student's evolving reasoning states, alleviating the off-policy issue. After training, the teacher performs better at continuing from student-generated prefixes, is less uncertain when conditioned on them, and therefore can provide more reliable supervision for the student.

\begin{figure*}[t]
    \centering

    % ==================== Left: control ====================
    \begin{minipage}[t]{0.47\textwidth}
        \vspace{0pt} % important
        \centering

        \captionof{table}{
        \textit{Student-conditioned teacher adaptation drives the gains.}
        OPD + Teacher GRPO applies the same outcome-reward teacher updates as \method, but starts teacher rollouts from the original problem rather than from student-generated prefixes.
        }
        \label{tab:ablation-prefix0}

        \small
        \setlength{\tabcolsep}{4pt}
        \begin{tabular}{lccc}
        \toprule
        \multirow{2}{*}{Method}
        & \multicolumn{2}{c}{Qwen3-4B}
        & \multicolumn{1}{c}{Qwen3-8B} \\
        \cmidrule(lr){2-3}\cmidrule(lr){4-4}
        & Math & Code & Math \\
        \midrule
        OPD
        & \underline{49.2}
        & 56.6
        & 49.0 \\
        OPD + Teacher GRPO
        & 48.8
        & \underline{57.0}
        & \underline{50.8} \\
        \midrule
        \textbf{SCOUT}
        & \textbf{51.4}
        & \textbf{59.7}
        & \textbf{51.6} \\
        \bottomrule
        \end{tabular}
    \end{minipage}
    \hfill
    % ==================== Right: update interval ====================
    \begin{minipage}[t]{0.50\textwidth}
        \vspace{0pt} % important
        \centering

        \includegraphics[width=0.85\linewidth]{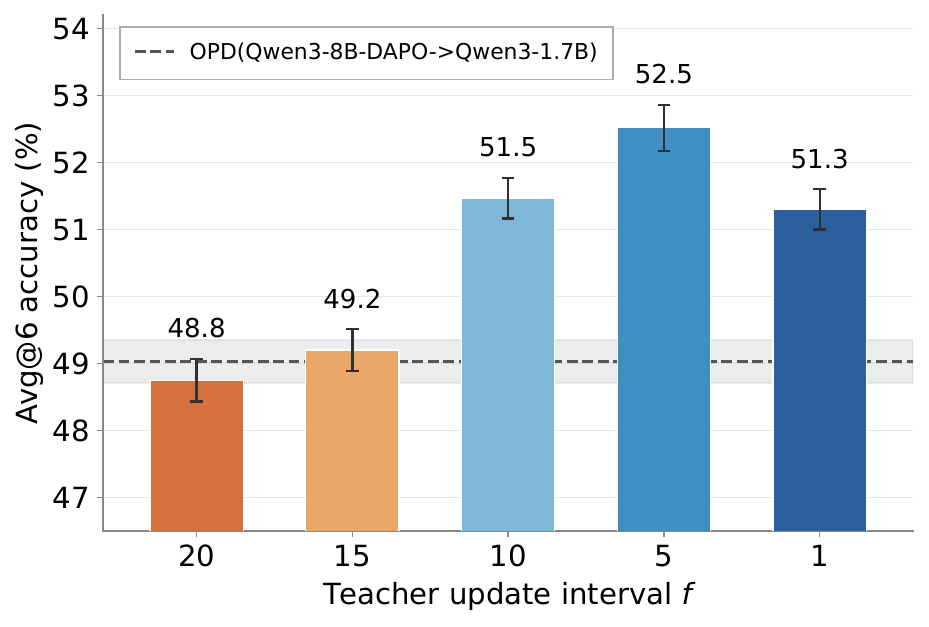}
        \vspace{-10pt}
        \captionof{figure}{
        \textit{Frequent teacher updates are unnecessary.}
        In the Qwen3-8B-DAPO $\rightarrow$ Qwen3-1.7B math setting, performance remains similar across moderate update intervals and drops only when updates become too sparse.
        }
        \label{fig:update_freq}
    \end{minipage}
\end{figure*}

\paragraph{\textcolor{deepblue}{(RQ2)} \method~gains come from adapting the teacher to student-generated prefixes.}
Relative to standard OPD, \method~optimizes the teacher during training \textit{and} conditions those updates on student-generated prefixes. This raises a natural question: do the gains come from additional teacher optimization, or from adapting the teacher specifically to student-generated states? To test this, we introduce \textbf{OPD + Teacher GRPO}. It trains the teacher with the same outcome-reward GRPO objective as \method, but on rollouts that start from the original problem. In contrast, \method~starts these rollouts from prefixes generated by the student. This control tests whether additional teacher training alone is sufficient, without conditioning on student-generated states. Table~\ref{tab:ablation-prefix0} compares OPD, OPD + Teacher GRPO, and \method~across two teacher sizes, and on both math and code. Full per-benchmark results and individual training runs are reported in Tables~\ref{tab:teacher_grpo_control_math_avg6} and~\ref{tab:teacher_grpo_control_code_4b}.

Additional teacher training alone does not match the gains of \method. On 4B math, OPD + Teacher GRPO slightly underperforms standard OPD, while on 4B code and 8B math it yields only modest improvements. In contrast, \method~achieves the strongest performance in all three settings. These results show that the gains do not come solely from additional teacher optimization; adapting the teacher to student-generated states is also important.

\paragraph{\textcolor{deepblue}{(RQ3)} \method~does not require frequent teacher updates.}
\label{sec:update-interval}
We vary the teacher update interval $f$ in the Qwen3-8B-DAPO $\rightarrow$ Qwen3-1.7B math setting to study how frequently the teacher must adapt as the student evolves. Figure~\ref{fig:update_freq} shows a broad range of effective update frequencies. Updating every 1, 5, or 10 student steps gives similar performance, with Avg@6 of $51.3$, $52.5$, and $51.5$, respectively. However, when updates become too sparse ($f{=}20$), performance falls close to standard OPD.

\begin{wrapfigure}[18]{r}{0.46\textwidth}
    \vspace{-10pt}
    \centering
    \includegraphics[width=0.9\linewidth]{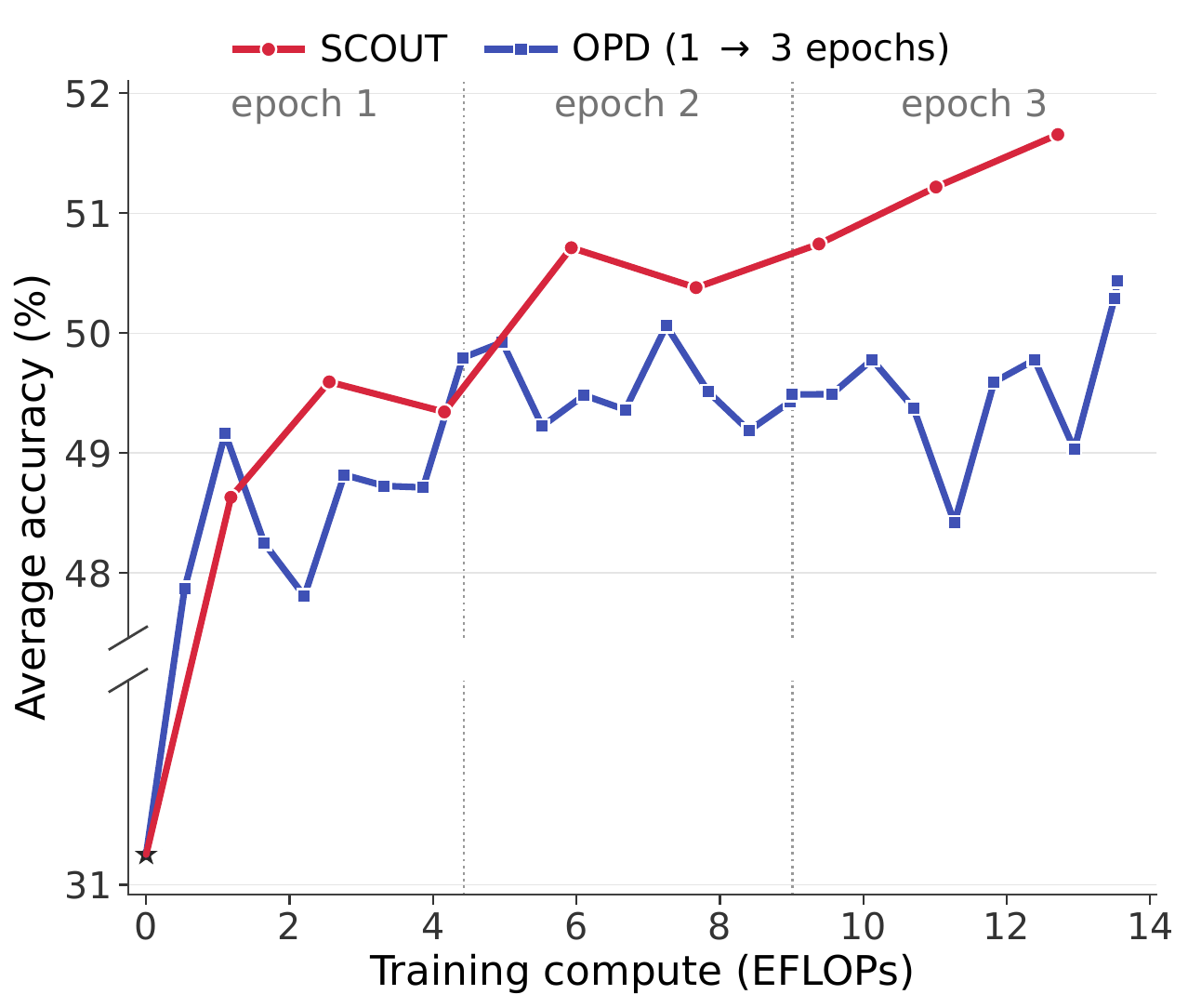}
    \caption{
\textit{Additional OPD training does not close the gap to \method.}
In the Qwen3-4B-Instruct-2507 $\rightarrow$ Qwen3-1.7B math setting, we train \method~for one epoch and OPD for three, on the same data, to compare performance over similar training-compute budgets. OPD quickly plateaus across repeated epochs, while \method~continues to improve.
    }
    \label{fig:flops}
\end{wrapfigure}

These results show that the teacher must adapt to the evolving student, but does not need to follow every student update. We therefore use $f=10$ in our main experiments: it preserves most of the performance gains while requiring fewer teacher updates. Full per-benchmark results are reported in Table~\ref{tab:analysis-update-freq}.

\textbf{Additional OPD training does not recover the gains of \method.}
Unlike standard OPD, \method~spends additional compute on periodic GRPO updates to the teacher. We therefore test whether standard OPD can recover these gains simply by training longer on the same data. In the Qwen3-4B-Instruct-2507 $\rightarrow$ Qwen3-1.7B setting, we evaluate both methods every 40 training steps. We train \method~for one epoch and OPD for three, giving the two methods comparable total training compute.

Figure~\ref{fig:flops} shows that OPD improves initially but quickly plateaus, whereas \method~continues to improve as training compute increases. Since the additional OPD epochs reuse the same training data, this comparison does not rule out further gains from fresh data. Within this fixed-data setting, however, additional OPD training does not close the gap to \method.

\paragraph{\method~complements loss-level improvements to OPD.}
\label{sec:optr-scout}
\method~adapts the teacher to student-generated prefixes, but this addresses only
one source of unreliable supervision in OPD. Loss-level methods instead modify
how the teacher signal is used during distillation. If these approaches address
different failure modes, their gains should be complementary. We test this by
combining \method~with On-Policy Trust Region (which we 
shorten as OPTR), the first component of Trust Region On-Policy Distillation (TrOPD)~\citep{xing2026trust}. We evaluate OPD, OPTR, \method, and OPTR+\method~in the Qwen3-4B-Instruct-2507 $\rightarrow$ Qwen3-1.7B setting.

\begin{wraptable}[11]{r}{0.5\textwidth}
    \centering
    \caption{
    \textit{\method~provides complementary gains on top of OPTR.}
    In the Qwen3-4B-Instruct-2507 $\rightarrow$ Qwen3-1.7B setting, adding \method~improves both OPD and OPTR across math and code.
}
    \vspace{-1em}
    \label{tab:optr-scout}
    \small
    \setlength{\tabcolsep}{3.5pt}
    \renewcommand{\arraystretch}{1.10}

    \begin{tabular}{llccc}
    \toprule
    \textbf{Task}
    & \textbf{Base}
    & \textbf{w/o \method}
    & \textbf{+ \method}
    & \textbf{Gain} \\
    \midrule

    \multirow{2}{*}{Math}
    & OPD
    & 49.2
    & \cellcolor{summarygray}\underline{51.4}
    & \cellcolor{summarygray}+2.2 \\

    & OPTR
    & 49.8
    & \cellcolor{scoutsummary}\textbf{52.3}
    & \cellcolor{scoutsummary}+2.5 \\

    \midrule

    \multirow{2}{*}{Code}
    & OPD
    & 56.6
    & \cellcolor{summarygray}\underline{59.7}
    & \cellcolor{summarygray}+3.1 \\

    & OPTR
    & 59.3
    & \cellcolor{scoutsummary}\textbf{60.5}
    & \cellcolor{scoutsummary}+1.2 \\

    \bottomrule
    \end{tabular}

\end{wraptable}

OPTR improves over standard OPD on both math and code, and combining it with \method~yields a further $2.5$-point gain on math and $1.2$-point gain on code, reaching the strongest performance in both settings. This shows that teacher adaptation remains beneficial even with an improved distillation objective, suggesting that the two approaches address complementary aspects of teacher--student mismatch. Detailed per-benchmark and per-run results are reported in Table~\ref{tab:optr_scout_math_full} and \ref{tab:optr_scout_code_full}.

%% file: sections_arxiv/5.5Relatedwork.tex
\section{Related Work}

OPD trains the student on trajectories sampled from its own policy while using a stronger teacher for token-level supervision \citep{agarwal2024onpolicydistillationlanguagemodels, gu2026minillmonpolicydistillationlarge}. As the student reasons, the teacher must supervise prefixes generated by the student rather than by its own policy, creating a teacher-side distribution shift. Much of recent work addresses this problem while keeping the teacher fixed and changing where or how its supervision is used.

One group of methods controls the rollout horizon. Prefix distillation and ESR stop supervision after an early prefix \citep{zhang2026fast,zhou2026less}. POPD gradually extends the rollout horizon during training, while TOPD uses a fixed truncated horizon \citep{zhang2026full}. Prune-OPD monitors local teacher--student compatibility and, when the models drift apart, down-weights later supervision and truncates the rollout~\citep{yang2026prune}.

Other methods retain more of the trajectory but change which teacher signals the student learns from. IW-OPD reweights tokens using the accumulated teacher--student discrepancy~\citep{xie2026positionbiasonpolicydistillation}, while SOD adjusts the distillation weight at each reasoning step using step-level divergence~\citep{zhong2026sod}. TIP identifies informative tokens using student entropy and teacher--student divergence~\citep{xu2026tip}. TrOPD restricts standard OPD to regions where teacher supervision is considered reliable and treats outlier regions separately~\citep{xing2026trust}. LGR instead looks one step ahead, favoring student tokens that lead to higher teacher confidence at the next step~\citep{liu2026teachercanthelphere}. Verifier-based methods incorporate task outcomes into the supervision signal: RG-OPD uses verifier feedback to gate teacher supervision \citep{akhondzadeh2026rewardgatedonpolicydistillation}, OPDVR uses trajectory correctness to gate token-level rewards \citep{lin2026onpolicydistillationverifiablereward}, and SPOT uses verifier-scored continuations to construct outcome-calibrated distillation targets~\citep{qu2026spotsparseprobingoutcome}.

A related line of work changes the trajectory or the form of corrective guidance. MOTAB detects student drift, backtracks to an earlier safe state, and uses the teacher to redirect generation \citep{wang2026backtrackingstraysmitigatingdual}. Relay-OPD lets the teacher briefly take over at detected failure points before returning generation to the student~\citep{xu2026relay}. TRD revises student rollouts under teacher guidance before distillation~\citep{jiang2026trajectory}. SOPD keeps student-visited prefixes but
asks the teacher to generate a complete reasoning step from each prefix, providing longer-horizon supervision than token-level OPD \citep{sun2026sopd}. TrOPD also includes an off-policy guidance component in which the student continues from teacher-generated prefixes \citep{xing2026trust}.

Across these approaches, the teacher itself remains fixed. They improve supervision by changing which states are visited, which teacher signals are used, or how those signals are presented to the student. \method~takes a different but complementary direction: it updates the teacher itself so that it becomes better at supervising student-generated states. Teacher adaptation therefore provides an additional optimization axis for OPD; our experiments further show that it can be combined with the loss-level OPTR component of TrOPD.

%% file: sections_arxiv/6Conclusion.tex
\section{Conclusions}
We introduced \method, a teacher--student co-training framework that addresses teacher-side distribution shift in on-policy distillation by adapting the teacher to student-generated prefixes. Across model pairs and task domains, \method~consistently improves over standard OPD and competing OPD methods. Our analysis reveals that the teacher becomes better at reasoning from student-generated prefixes, and that the gains come primarily from adapting it to the student rather than simply training the teacher further. Together, these results establish teacher adaptation as a promising and complementary optimization axis for improving on-policy distillation.

\section*{Acknowledgements}
The authors would like to thank Amir Tahmasbi, Zhehui Huang, Karen Hovsepian, and Zhishen Huang for helpful feedback and discussions throughout the project. We also thank Vijay Lingam for early discussion on OPD.

\clearpage

%% file: sections_arxiv/7Appendix.tex
\section{Limitations and Future Work}
Following recent OPD work, our experiments focus on mathematical reasoning and function-level code generation. We broaden this evaluation across multiple teacher--student pairs, model families, and task domains. However, these benchmarks still represent relatively contained reasoning problems rather than long-horizon agentic tasks. Evaluating \method~on settings where trajectories extend over many decisions, such as repository-level software engineering or interactive agents, is therefore an important direction for future work. \method~also introduces additional cost by periodically generating teacher rollouts and updating the teacher with RL. Although our compute-matched analysis shows that simply training OPD longer on the same data does not recover the gains, \method~still incurs additional wall-clock and memory costs. Developing more efficient teacher updates is therefore an important next step. Moreover, our current teacher adaptation relies on verifiable outcome rewards. We therefore see learned or weaker feedback as a natural path toward extending \method~beyond verifiable domains. Finally, \method~assumes that the teacher model is directly accessible and trainable, and is thus not immediately applicable to black-box or API-based teachers whose parameters cannot be updated. One possible direction is to replace parameter updates with prompt-based teacher adaptation, potentially extending \method~to settings with inaccessible teachers.

\section{Experiment Details}
\label{app:A}
\subsection{Hyperparameter Settings}
\label{app:hyperparameter}
We present the general hyperparameter settings in Tables~\ref{tab:hyper-math} and ~\ref{tab:hyper-code}. The performance varies with the choice of learning rate. Therefore, we conduct a grid search for the student model learning rate among $\{1, 3, 5, 7\}\times 10^{-6}$ on standard \textbf{OPD}. With the fixed student learning rate, we further search the teacher model learning rate among $\{1, 2, 3, 4, 5\}\times 10^{-6}$ for \method. The best learning rates are shown in Table~\ref{tab:learning_rates}.

\textbf{GRPO} uses a learning rate of $3\times 10^{-6}$. The rest of the hyperparameters are the same as OPD, except that its \textit{Rollout Number per Prompt (n)} is 8. The reference weight $\beta=0.001$.

In \method, the update of the teacher model uses GRPO with the same settings as student-side GRPO described above, except for its learning rate shown in Table~\ref{tab:learning_rates}.

\begin{table}[h]
\centering
\begin{minipage}[t]{0.48\linewidth}
\centering
\caption{Hyperparameters in Math reasoning task.}
\label{tab:hyper-math}
\begin{tabular}{lc}
\toprule
\textbf{Hyperparameter} & \textbf{Value} \\
\midrule
Max Prompt Length               & 1024 \\
Max Response Length             & 16384 \\
Training Rollout Temperature             & 1.0 \\
Training Rollout Top-$p$                 & 1.0 \\
Rollout Number per Prompt ($n$) & 1 \\
Global Batch Size               & 128 \\
PPO Mini-Batch Size             & 128 \\
PPO Epochs                      & 1 \\
PPO Clipping Range              & 0.2 \\
Learning-Rate Schedule          & constant \\
Training Epochs                 & 1 \\
\midrule
Inference Temperature & 0.6 \\
Inference Top-$p$ & 0.95 \\
Inference Max Response Length & 16384 \\
\bottomrule
\end{tabular}
\end{minipage}
\hfill
\begin{minipage}[t]{0.48\linewidth}
\centering
\caption{Hyperparameters in Code generation task.}
\label{tab:hyper-code}
\begin{tabular}{lc}
\toprule
\textbf{Hyperparameter} & \textbf{Value} \\
\midrule
Max Prompt Length               & 1536 \\
Max Response Length             & 16384 \\
Training Rollout Temperature             & 1.0 \\
Training Rollout Top-$p$                 & 1.0 \\
Rollout Number per Prompt ($n$) & 1 \\
Global Batch Size               & 128 \\
PPO Mini-Batch Size             & 128 \\
PPO Epochs                      & 1 \\
PPO Clipping Range              & 0.2 \\
Learning-Rate Schedule          & constant \\
Training Epochs                 & 3 \\
\midrule
Inference Temperature & 0.6 \\
Inference Top-$p$ & 0.95 \\
Inference Max Response Length & 16384 \\
\bottomrule
\end{tabular}
\end{minipage}
\end{table}

\begin{table}[t]
\centering
\small
\caption{Learning rates for the student and teacher models across different model pairs and tasks.}
\label{tab:learning_rates}
\begin{tabular}{lllcc}
\toprule
Teacher Model & Student Model & Task & Student LR & Teacher LR \\
\midrule
Qwen3-4B-Instruct
& Qwen3-1.7B
& Math & $5{\times}10^{-6}$ & $5{\times}10^{-6}$ \\

Qwen3-8B-DAPO
& Qwen3-1.7B
& Math & $3{\times}10^{-6}$ & $5{\times}10^{-6}$ \\

Skywork-OR1-MATH-7B
& DeepSeek-R1-Distill-Qwen-1.5B
& Math & $3{\times}10^{-6}$ & $5{\times}10^{-6}$ \\

Qwen3-4B-Instruct
& Qwen3-1.7B
& Code & $5{\times}10^{-6}$ & $5{\times}10^{-6}$ \\

\bottomrule
\end{tabular}
\end{table}

\subsection{Loss Implementation}

Although OPD is formulated using the reverse KL divergence, computing the exact full-vocabulary divergence requires evaluating the teacher distribution over the entire vocabulary at every student-generated token, which introduces substantial computational and memory overhead. We therefore adopt the policy-gradient formulation of OPD and use the single-sample K1 estimator. Specifically, for each student-generated token $y_t$, we define the token-level advantage as
\begin{equation}
A_t^{\mathrm{OPD}}
=
\operatorname{sg}\!\left[
\log \pi_T(y_t \mid s_t)
-
\log \pi_{\theta_{\mathrm{old}}}(y_t \mid s_t)
\right],
\end{equation}
where $\operatorname{sg}[\cdot]$ denotes the stop-gradient operator and $\pi_{\theta_{\mathrm{old}}}$ is the student policy used to generate the rollout. The student is then optimized with the clipped policy-gradient objective
\begin{equation}
\mathcal{L}_{\mathrm{OPD}}(\theta)
=
-
\mathbb{E}_{t}
\left[
\min\left(
r_t(\theta) A_t^{\mathrm{OPD}},
\operatorname{clip}\!\left(
r_t(\theta),
1-\epsilon_{\mathrm{low}},
1+\epsilon_{\mathrm{high}}
\right)
A_t^{\mathrm{OPD}}
\right)
\right],
\end{equation}
where
\begin{equation}
r_t(\theta)
=
\frac{
\pi_{\theta}(y_t \mid s_t)
}{
\pi_{\theta_{\mathrm{old}}}(y_t \mid s_t)
}
\end{equation}
is the importance-sampling ratio between the updated and rollout policies. This formulation avoids explicitly computing the full-vocabulary KL divergence while providing an efficient sampled approximation to the reverse-KL objective.

\subsection{Evaluation Details}
\label{app:eval-details}
We adopt 6 benchmarks for the reasoning task and 3 benchmarks for the code generation task. Table~\ref{tab:benchmark-intro} lists the number of questions for each benchmark and the number of repeated evaluations we conduct. We repeat 32 times for benchmarks with few questions: AIME 24, AIME 25, AMC, and HMMT. For HumanEval+, we evaluate them 16 times. For MATH500 and MBPP, we evaluate them 8 times. For OlympiadBench, we use the text-only English subset, "OE\_TO\_maths\_en\_COMP", and filter the single-answer questions, resulting in 581 questions. For LiveCodeBench, we use the stable v5 version with 880 questions, and repeat evaluation 4 times.

For each OPD-based method, we conduct three independent training runs and
evaluate every resulting checkpoint using the repeated-evaluation protocol
described above.
Tables~\ref{tab:math_4b_to_1p7b_per_run},
\ref{tab:code_4b_to_1p7b_per_run},
\ref{tab:math_8b_to_1p7b_per_run}, and
\ref{tab:math_skywork_to_1p7b_per_run} provide the detailed results.
For an individual training run, we report the mean accuracy across repeated
evaluations together with its standard error of the mean (SEM). In the
aggregate rows, we report the mean across the three training runs together
with the sample standard deviation (SD) of their run-level mean accuracies.

More precisely, let $y_{m,b,r}$ and $e_{m,b,r}$ denote the mean accuracy and
within-evaluation SEM, respectively, for method $m$, benchmark $b$, and
training run $r$. With $R=3$ training runs, the aggregate accuracy and the
reported cross-run SD are
\begin{equation}
\bar{y}_{m,b}
=
\frac{1}{R}\sum_{r=1}^{R}y_{m,b,r},
\qquad
s_{m,b}^{2}
=
\frac{1}{R-1}
\sum_{r=1}^{R}
\left(y_{m,b,r}-\bar{y}_{m,b}\right)^{2}.
\end{equation}
Thus, the aggregate table entry is $\bar{y}_{m,b}\pm s_{m,b}$.
For significance testing on an individual benchmark, we use a two-level
random-effects variance decomposition that separates variation across
training runs from uncertainty due to repeated evaluation. We estimate the
between-run variance as
\begin{equation}
\hat{\tau}_{m,b}^{2}
=
\max\left(
0,\,
s_{m,b}^{2}
-
\frac{1}{R}\sum_{r=1}^{R}e_{m,b,r}^{2}
\right),
\end{equation}
and estimate the variance of the aggregate mean by
\begin{equation}
V_{m,b}
=
\frac{\hat{\tau}_{m,b}^{2}}{R}
+
\frac{1}{R^{2}}\sum_{r=1}^{R}e_{m,b,r}^{2}.
\end{equation}
For a method $m$ and the corresponding OPD baseline $0$ (Frozen-OPD in the
Skywork setting), we test
$H_0:\mu_{m,b}\leq\mu_{0,b}$ against
$H_1:\mu_{m,b}>\mu_{0,b}$ using
\begin{equation}
z
=
\frac{\bar{y}_{m,b}-\bar{y}_{0,b}}
{\sqrt{V_{m,b}+V_{0,b}}},
\qquad
p=1-\Phi(z),
\end{equation}
where $\Phi$ is the standard normal cumulative distribution function.

For the Avg. column, we first compute the equal-weight macro-average within
each training run,
\begin{equation}
M_{m,r}
=
\frac{1}{B}\sum_{b=1}^{B}y_{m,b,r},
\end{equation}
and then aggregate these run-level macro-averages. Here, $B=6$ for the mathematical reasoning results, and $B=3$ for code generation. We compare the three run-level macro-averages of each method against those of
the baseline using a one-sided Welch's $t$-test:
\begin{equation}
t
=
\frac{\bar{M}_{m}-\bar{M}_{0}}
{\sqrt{s_{M,m}^{2}/R_m+s_{M,0}^{2}/R_0}},
\end{equation}
with Welch--Satterthwaite degrees of freedom
\begin{equation}
\nu
=
\frac{
\left(s_{M,m}^{2}/R_m+s_{M,0}^{2}/R_0\right)^{2}
}{
\frac{\left(s_{M,m}^{2}/R_m\right)^{2}}{R_m-1}
+
\frac{\left(s_{M,0}^{2}/R_0\right)^{2}}{R_0-1}
}.
\end{equation}
The one-sided $p$-value is $p=1-F_{t_{\nu}}(t)$. Orange cells indicate cases
in which a method achieves a higher mean than the corresponding baseline and
the one-sided test gives $p<0.1$. The main experimental tables report the
corresponding aggregate mean and cross-run SD over the three training runs.

\begin{table}[t]
\centering
\caption{The statistics of the evaluation benchmarks, including the number of questions and the number of sampled responses per question.}
\label{tab:benchmark-intro}
\setlength{\tabcolsep}{4pt}
\renewcommand{\arraystretch}{1.15}
\resizebox{\columnwidth}{!}{%
\begin{tabular}{@{}l ccccccc ccc@{}}
\toprule
\multirow{2}{*}{\textbf{Benchmark}}
  & \multicolumn{6}{c}{\textbf{Reasoning}}
  & \multicolumn{3}{c}{\textbf{Code Generation}} \\
\cmidrule(lr){2-7} \cmidrule(lr){8-10}
  & AIME24 & AIME25 & AMC23 & HMMT & Olympiad & MATH500
  & LCB v5 & HumanEval+ & MBPP \\
\midrule
\# Questions & 30 & 30 & 40 & 30 & 581 & 500 & 880 & 164 & 500 \\
\# Repeats   & 32 & 32 & 32 & 32 & 4   & 8   & 4   & 16   & 8   \\
\bottomrule
\end{tabular}%
}
\end{table}

\subsection{Reproducibility and Compute}

All experiments are run on a Slurm-managed cluster of AWS p4d.24xlarge
instances, each equipped with 8 NVIDIA A100 40GB GPUs. All trainable methods are run with three independent seeds.
We use no SFT warm-up stage: training starts directly from the released
instruction or reasoning checkpoints.

For the Qwen3-8B-DAPO teacher used in our math experiments, we train
Qwen3-8B with critic-free GRPO on the deduplicated DAPO-Math-17k dataset for exactly three epochs. We use a train and mini-batch size of
64, 8 rollouts per prompt, a constant learning rate of $1\times10^{-6}$, rollout temperature 1.0. 

For math experiments, GRPO, Prune-OPD, and Relay-OPD use one node
(8 GPUs), while frozen-teacher OPD and ESR use two nodes (16 GPUs).
SCOUT uses three nodes (24 GPUs) with a 4B teacher and four nodes
(32 GPUs) with a 7B/8B teacher. Code experiments use the same general
resource allocation. In \method, student and teacher training are placed on
separate physical nodes using Ray resource pools. All evaluations are run
on a single 8-GPU node with independent vLLM workers.

Our main software stack uses PyTorch 2.9.0, vLLM 0.12.0, Ray 2.55.1,
verl 0.9.0.dev0, and Transformers 4.57.6. Training and inference
hyperparameters are provided in the preceding appendix sections.

\section{Complete Experiment Results}
\label{app:full-results}
This section provides the complete results corresponding to the experiments
reported in the main paper. In addition to the aggregate results presented in the main text, we report per-benchmark performance and individual training runs for each experimental setting.

\subsection{Results of the Main Experiments}
For the main experiments, each method is trained with three independent runs, unless otherwise specified.
For individual runs, the $\pm$ term denotes the standard error from repeated
evaluation, while in the \textit{Mean} rows it denotes the standard deviation
across training runs. One-sided $p$-values for improvement over OPD are computed as described in
Appendix~\ref{app:eval-details}; orange cells indicate $p<0.1$. Best and second-best three-run aggregate results are shown in
\textbf{bold} and \underline{underline}, respectively.

Tables~\ref{tab:math_4b_to_1p7b_per_run} and \ref{tab:code_4b_to_1p7b_per_run} correspond to the Qwen3-4B-Instruct-2507 $\rightarrow$ Qwen3-1.7B setting in Table~\ref{tab:qwen3-4b-main-results}. Relay-OPD fails the code generation task, as it generates excessive syntax errors. Therefore, we skipped repeated training on it. Table~\ref{tab:math_8b_to_1p7b_per_run} corresponds to the Qwen3-8B-DAPO $\rightarrow$ Qwen3-1.7B setting in Table~\ref{tab:qwen3-8b-math-results}, and Table~\ref{tab:math_skywork_to_1p7b_per_run} corresponds to the Skywork-OR1-Math-7B $\rightarrow$ DeepSeek-R1-Distill-Qwen-1.5B setting in Table~\ref{tab:deepseek-math-results}.

\subsection{Results of the Analysis Experiments}
We present the complete per-benchmark results of the analysis experiments.
Tables~\ref{tab:teacher_grpo_control_math_avg6} and \ref{tab:teacher_grpo_control_code_4b} correspond to the ablation study on teacher GRPO conditions in Table~\ref{tab:ablation-prefix0}. Tables~\ref{tab:optr_scout_math_full} and \ref{tab:optr_scout_code_full} correspond to the analysis on whether \method~can complement other OPD methods in Table~\ref{tab:optr-scout}.

Table~\ref{tab:analysis-update-freq} corresponds to the comparison of different teacher update frequencies in Figure~\ref{fig:update_freq}. 
Importantly, we find that the teacher learning rate $5\times 10^{-6}$ is too large for frequent updates, and reduce it proportionally with $f$, to $2.5\times 10^{-6}$ and $5\times 10^{-7}$ for $f{=}5$ and $f{=}1$, respectively. Since $f{=}1$ is time-consuming, we skip the repeat run for this experiment and report the single-run result.

\input{Tables/table_4b_math_full}
\input{Tables/table_4b_code_full}
\input{Tables/table_8b_math_full}
\input{Tables/table_skywork_7b_math_full}

\input{Tables/table_prefix0_math_full}
\input{Tables/table_prefix0_code_full}
\input{Tables/table_ablation_f_8b}
\input{Tables/table_optr_scout_math_full}
\input{Tables/table_optr_scout_code_full}

%% file: Tables/table_4b_math_full.tex
\begin{table*}[t]
\centering
\caption{
Complete results for the Qwen3-4B-Instruct-2507 $\rightarrow$ Qwen3-1.7B setting on six math reasoning benchmarks. This table corresponds to the mathematical reasoning task in Table~\ref{tab:qwen3-4b-main-results}. \method~shows significant improvement over OPD on 3 of 6 benchmarks and in the average performance.
}
\label{tab:math_4b_to_1p7b_per_run}

\small
\setlength{\tabcolsep}{4.0pt}
\renewcommand{\arraystretch}{1.10}

\resizebox{\textwidth}{!}{
\begin{tabular}{llccccccc}
\toprule
\textbf{Method}
& \textbf{Run}
& \textbf{AIME24}
& \textbf{AIME25}
& \textbf{AMC23}
& \textbf{HMMT}
& \textbf{Olymp.}
& \textbf{MATH500}
& \textbf{Avg6} \\[-1pt]

&
\textit{Avg.@}
& \textit{32}
& \textit{32}
& \textit{32}
& \textit{32}
& \textit{4}
& \textit{8}
& -- \\

\midrule

% ============================================================
% GRPO
% ============================================================

\multirow{3}{*}{GRPO}
& Run 1
& \res{29.06}{0.97}
& \res{30.10}{0.61}
& \res{73.28}{0.76}
& \res{16.77}{0.69}
& \res{60.80}{0.48}
& \res{87.85}{0.11}
& $49.64$ \\

& Run 2
& \res{27.71}{0.98}
& \res{26.46}{0.70}
& \res{67.58}{0.84}
& \res{16.98}{0.48}
& \res{55.46}{0.22}
& \res{83.67}{0.42}
& $46.31$ \\

& Run 3
& \res{36.04}{1.26}
& \res{34.58}{0.87}
& \res{79.06}{0.94}
& \res{22.81}{0.54}
& \res{67.00}{0.33}
& \res{90.62}{0.26}
& $55.02$ \\

\rowcolor{summarygray}
& Mean
& \res{30.94}{4.47}
& \res{30.38}{4.07}
& \res{73.31}{5.74}
& \res{18.85}{3.43}
& \res{61.09}{5.78}
& \res{87.38}{3.50}
& \res{50.32}{4.39} \\

\midrule

% ============================================================
% OPD
% ============================================================

\multirow{3}{*}{OPD}
& Run 1
& \res{30.94}{1.06}
& \res{22.40}{0.81}
& \res{71.95}{0.71}
& \res{15.42}{0.65}
& \res{58.30}{0.45}
& \res{85.75}{0.52}
& $47.46$ \\

& Run 2
& \res{36.98}{0.97}
& \res{25.83}{0.87}
& \res{74.22}{0.82}
& \res{15.83}{0.54}
& \res{60.71}{0.31}
& \res{87.95}{0.40}
& $50.25$ \\

& Run 3
& \res{33.54}{1.00}
& \res{24.69}{0.86}
& \res{75.62}{0.84}
& \res{16.88}{0.86}
& \res{61.45}{0.60}
& \res{87.40}{0.19}
& $49.93$ \\

\rowcolor{summarygray}
& Mean
& \res{33.82}{3.03}
& \res{24.31}{1.75}
& \secondres{73.93}{1.85}
& \res{16.04}{0.75}
& \secondres{60.15}{1.64}
& \secondres{87.03}{1.14}
& \res{49.21}{1.53} \\

\midrule

% ============================================================
% ESR(4096)
% ============================================================

\multirow{3}{*}{ESR}
& Run 1
& \res{31.56}{1.31}
& \res{26.77}{0.71}
& \res{71.09}{0.74}
& \res{17.19}{0.43}
& \res{58.48}{0.28}
& \res{86.95}{0.15}
& $48.67$ \\

& Run 2
& \res{31.25}{1.16}
& \res{24.90}{0.78}
& \res{67.81}{0.66}
& \res{16.46}{0.67}
& \res{59.34}{0.43}
& \res{86.22}{0.25}
& $47.66$ \\

& Run 3
& \res{32.92}{1.11}
& \res{24.90}{0.97}
& \res{72.89}{0.79}
& \res{14.48}{0.71}
& \res{59.25}{0.45}
& \res{85.72}{0.26}
& $48.36$ \\

\rowcolor{summarygray}
& Mean
& \res{31.91}{0.89}
& \res{25.52}{1.08}
& \res{70.60}{2.58}
& \res{16.04}{1.40}
& \res{59.02}{0.47}
& \res{86.30}{0.62}
& \res{48.23}{0.52} \\

& \textit{$p$ vs.\ OPD}
& $0.803$
& $0.187$
& $0.925$
& $0.500$
& $0.821$
& $0.801$
& $0.808$ \\

\midrule

% ============================================================
% Prune-OPD
% ============================================================

\multirow{3}{*}{Prune-OPD}
& Run 1
& \res{33.44}{1.07}
& \res{27.60}{0.78}
& \res{68.98}{0.91}
& \res{15.73}{0.74}
& \res{59.51}{0.32}
& \res{86.17}{0.39}
& $48.57$ \\

& Run 2
& \res{33.65}{1.28}
& \res{27.92}{0.77}
& \res{70.00}{1.05}
& \res{16.56}{0.63}
& \res{57.87}{0.29}
& \res{85.72}{0.59}
& $48.62$ \\

& Run 3
& \res{34.90}{0.95}
& \res{26.46}{0.81}
& \res{70.39}{1.07}
& \res{17.29}{0.62}
& \res{58.61}{0.47}
& \res{85.85}{0.36}
& $48.92$ \\

\rowcolor{summarygray}
& Mean
& \secondres{33.99}{0.79}
& \secondres{27.33}{0.77}
& \res{69.79}{0.73}
& \secondres{16.53}{0.78}
& \res{58.66}{0.82}
& \res{85.91}{0.23}
& \res{48.70}{0.19} \\

& \textit{$p$ vs.\ OPD}
& $0.466$
& \cellcolor[RGB]{255,230,180}$0.0391$
& $0.980$
& $0.242$
& $0.871$
& $0.887$
& $0.689$ \\

\midrule

% ============================================================
% Relay-OPD
% ============================================================

\multirow{3}{*}{Relay-OPD}
& Run 1
& \res{37.19}{0.96}
& \res{23.54}{0.96}
& \res{73.05}{0.94}
& \res{13.23}{0.59}
& \res{60.07}{0.16}
& \res{86.10}{0.43}
& $48.86$ \\

& Run 2
& \res{37.60}{1.03}
& \res{25.00}{0.70}
& \res{75.23}{0.89}
& \res{15.73}{0.71}
& \res{59.68}{0.72}
& \res{86.30}{0.29}
& $49.92$ \\

& Run 3
& \res{32.50}{0.82}
& \res{25.94}{0.96}
& \res{73.44}{0.85}
& \res{15.94}{0.61}
& \res{59.81}{0.65}
& \res{86.42}{0.25}
& $49.01$ \\

\rowcolor{summarygray}
& Mean
& \bestres{35.76}{2.83}
& \res{24.83}{1.21}
& \res{73.91}{1.16}
& \res{14.97}{1.51}
& \res{59.85}{0.20}
& \res{86.27}{0.16}
& \secondres{49.27}{0.57} \\

& \textit{$p$ vs.\ OPD}
& $0.231$
& $0.348$
& $0.507$
& $0.825$
& $0.606$
& $0.815$
& $0.481$ \\

\midrule

% ============================================================
% SCOUT
% Selected runs: Ref (s0), Seed 2, Seed 3
% ============================================================

\multirow{3}{*}{\textbf{SCOUT}}
& Run 1
& \res{36.35}{0.94}
& \res{28.12}{0.73}
& \res{76.17}{0.65}
& \res{18.75}{0.55}
& \res{63.25}{0.66}
& \res{88.67}{0.27}
& $51.89$ \\

& Run 2
& \res{35.52}{1.06}
& \res{28.96}{0.87}
& \res{75.08}{0.71}
& \res{15.10}{0.54}
& \res{61.57}{0.41}
& \res{88.38}{0.24}
& $50.77$ \\

& Run 3
& \res{35.42}{1.03}
& \res{29.38}{0.84}
& \res{76.02}{0.78}
& \res{17.81}{0.71}
& \res{62.31}{0.35}
& \res{88.17}{0.29}
& $51.52$ \\

\rowcolor{scoutsummary}
& Mean
& \bestres{35.76}{0.51}
& \bestres{28.82}{0.64}
& \bestres{75.76}{0.59}
& \bestres{17.22}{1.90}
& \bestres{62.38}{0.84}
& \bestres{88.41}{0.25}
& \bestres{51.39}{0.57} \\

& \textit{$p$ vs.\ OPD}
& $0.192$
& \cellcolor[RGB]{255,230,180}$0.0151$
& $0.112$
& $0.200$
& \cellcolor[RGB]{255,230,180}$0.0647$
& \cellcolor[RGB]{255,230,180}$0.0839$
& \cellcolor[RGB]{255,230,180}$0.0598$ \\

\bottomrule
\end{tabular}
}
\end{table*}

%% file: Tables/table_4b_code_full.tex
\begin{table*}[t]
\centering
\caption{
Per-run results on coding benchmarks for the
Qwen3-4B-Instruct-2507 $\rightarrow$ Qwen3-1.7B setting.
This table corresponds to the code generation task in Table~\ref{tab:qwen3-4b-main-results}. While other OPD-based methods perform worse in the code domain, \method~demonstrates its generalizability across domains, with significantly better accuracy on 2 of 3 benchmarks and in the average performance.
}
\label{tab:code_4b_to_1p7b_per_run}

\small
\setlength{\tabcolsep}{7.0pt}
\renewcommand{\arraystretch}{1.10}

\resizebox{\textwidth}{!}{
\begin{tabular}{llcccc}
\toprule
\textbf{Method}
& \textbf{Run}
& \textbf{LCB v5}
& \textbf{HumanEval+}
& \textbf{MBPP}
& \textbf{Avg.} \\[-1pt]

&
\textit{Avg.@}
& \textit{4}
& \textit{16}
& \textit{8}
& -- \\

\midrule

% ============================================================
% GRPO -- single run
% ============================================================

GRPO
& Run 1
& \res{43.12}{0.65}
& \res{74.16}{1.64}
& \res{64.20}{0.85}
& $60.49$ \\
& Run 2
& \res{38.81}{1.50}
& \res{72.71}{2.94}
& \res{70.85}{1.63}
& $60.79$ \\
& Run 3
& \res{51.02}{1.53}
& \res{71.34}{2.73}
& \res{73.88}{1.57}
& $65.41$ \\
\rowcolor{summarygray}
& Mean
& \res{44.32}{6.20}
& \res{72.74}{1.41}
& \res{69.64}{4.95}
& \res{62.23}{2.76} \\

\midrule

% ============================================================
% OPD
% ============================================================

\multirow{3}{*}{OPD}
& Run 1
& \res{37.67}{0.25}
& \res{69.21}{0.82}
& \res{62.50}{0.36}
& $56.46$ \\

& Run 2
& \res{37.67}{0.55}
& \res{69.51}{0.78}
& \res{62.18}{0.42}
& $56.45$ \\

& Run 3
& \res{37.59}{0.29}
& \res{70.50}{0.49}
& \res{62.75}{0.34}
& $56.95$ \\

\rowcolor{summarygray}
& Mean
& \secondres{37.64}{0.05}
& \secondres{69.74}{0.68}
& \secondres{62.48}{0.29}
& \secondres{56.62}{0.28} \\

\midrule

% ============================================================
% ESR(4096)
% ============================================================

\multirow{3}{*}{ESR}
& Run 1
& \res{37.05}{0.51}
& \res{69.05}{0.61}
& \res{61.85}{0.31}
& $55.98$ \\

& Run 2
& \res{33.41}{0.27}
& \res{67.15}{0.87}
& \res{58.80}{0.37}
& $53.12$ \\

& Run 3
& \res{37.56}{0.36}
& \res{68.06}{0.85}
& \res{63.10}{0.51}
& $56.24$ \\

\rowcolor{summarygray}
& Mean
& \res{36.00}{2.26}
& \res{68.09}{0.95}
& \res{61.25}{2.21}
& \res{55.11}{1.73} \\

& \textit{$p$ vs.\ OPD}
& $0.832$
& $0.961$
& $0.781$
& $0.865$ \\

\midrule

% ============================================================
% Prune-OPD
% ============================================================

\multirow{3}{*}{Prune-OPD}
& Run 1
& \res{36.73}{0.22}
& \res{68.94}{0.73}
& \res{60.80}{0.27}
& $55.49$ \\

& Run 2
& \res{36.70}{0.59}
& \res{69.13}{0.70}
& \res{61.30}{0.40}
& $55.71$ \\

& Run 3
& \res{37.05}{0.64}
& \res{70.16}{0.46}
& \res{61.52}{0.58}
& $56.24$ \\

\rowcolor{summarygray}
& Mean
& \res{36.83}{0.19}
& \res{69.41}{0.66}
& \res{61.21}{0.37}
& \res{55.81}{0.39} \\

& \textit{$p$ vs.\ OPD}
& $0.950$
& $0.707$
& $0.990$
& $0.976$ \\

\midrule

% ============================================================
% Relay-OPD -- Run 2/3 placeholders
% ============================================================

\multirow{3}{*}{Relay-OPD}
& Run 1
& \res{2.3}{0.4}
& \res{14.3}{1.6}
& \res{19.6}{0.9}
& $12.07$ \\

& Run 2
& -- & -- & -- & -- \\

& Run 3
& -- & -- & -- & -- \\

\midrule

% ============================================================
% SCOUT
% ============================================================

% \multirow{3}{*}{\textbf{SCOUT}}
% & Run 1
% & \res{39.97}{0.34}
% & \res{73.93}{0.73}
% & \res{66.70}{0.61}
% & $60.20$ \\

% & Run 2
% & \res{43.32}{0.44}
% & \res{73.93}{0.36}
% & \res{66.77}{0.39}
% & $61.34$ \\

% & Run 3
% & \res{38.41}{0.40}
% & \res{72.41}{0.56}
% & \res{62.20}{0.38}
% & $57.67$ \\

% \rowcolor{scoutsummary}
% & Mean
% & \bestres{40.57}{2.51}
% & \bestres{73.42}{0.88}
% & \bestres{65.22}{2.62}
% & \bestres{59.74}{1.88} \\

% & \textit{$p$ vs.\ OPD}
% & $0.272$
% & $0.138$
% & $0.179$
% & $0.162$ \\

% \bottomrule
\multirow{3}{*}{\textbf{SCOUT}}
& Run 1
& \res{39.97}{0.34}
& \res{73.93}{0.73}
& \res{66.70}{0.61}
& $60.20$ \\
 
& Run 2
& \res{43.32}{0.44}
& \res{73.93}{0.36}
& \res{66.77}{0.39}
& $61.34$ \\
 
& Run 3
& \res{38.41}{0.40}
& \res{72.41}{0.56}
& \res{62.20}{0.38}
& $57.67$ \\
 
\rowcolor{scoutsummary}
& Mean
& \bestres{40.57}{2.51}
& \bestres{73.42}{0.88}
& \bestres{65.22}{2.62}
& \bestres{59.74}{1.88} \\
 
& \textit{$p$ vs.\ OPD}
& \cellcolor[RGB]{255,230,180}$0.0905$
& \cellcolor[RGB]{255,230,180}$0.0027$
& $0.105$
& \cellcolor[RGB]{255,230,180}$0.0497$ \\
 
\bottomrule
\end{tabular}
}
\end{table*}

%% file: Tables/table_8b_math_full.tex
\begin{table*}[t]
\centering
\caption{
Complete per-run results for the Qwen3-8B-DAPO
$\rightarrow$ Qwen3-1.7B setting on six math reasoning benchmarks.
This table corresponds to Table~\ref{tab:qwen3-8b-math-results}. \method~significantly outperforms the OPD baseline in 5 of 6 benchmarks and in average performance. These results suggest that \method’s gains extend to the larger-teacher setting.
}
\label{tab:math_8b_to_1p7b_per_run}

\small
\setlength{\tabcolsep}{4.0pt}
\renewcommand{\arraystretch}{1.10}

\resizebox{\textwidth}{!}{
\begin{tabular}{llccccccc}
\toprule
\textbf{Method}
& \textbf{Run}
& \textbf{AIME24}
& \textbf{AIME25}
& \textbf{AMC23}
& \textbf{HMMT}
& \textbf{Olymp.}
& \textbf{MATH500}
& \textbf{Avg6} \\[-1pt]

&
\textit{Avg.@}
& \textit{32}
& \textit{32}
& \textit{32}
& \textit{32}
& \textit{4}
& \textit{8}
& -- \\

\midrule

% ============================================================
% GRPO
% ============================================================

\multirow{3}{*}{GRPO}
& Run 1
& \res{29.06}{0.97}
& \res{30.10}{0.61}
& \res{73.28}{0.76}
& \res{16.77}{0.69}
& \res{60.80}{0.48}
& \res{87.85}{0.11}
& $49.64$ \\

& Run 2
& \res{27.71}{0.98}
& \res{26.46}{0.70}
& \res{67.58}{0.84}
& \res{16.98}{0.48}
& \res{55.46}{0.22}
& \res{83.67}{0.42}
& $46.31$ \\

& Run 3
& \res{36.04}{1.26}
& \res{34.58}{0.87}
& \res{79.06}{0.94}
& \res{22.81}{0.54}
& \res{67.00}{0.33}
& \res{90.62}{0.26}
& $55.02$ \\

\rowcolor{summarygray}
& Mean
& \res{30.94}{4.47}
& \res{30.38}{4.07}
& \res{73.31}{5.74}
& \res{18.85}{3.43}
& \res{61.09}{5.78}
& \res{87.38}{3.50}
& \res{50.32}{4.39} \\

\midrule

% ============================================================
% OPD
% ============================================================

\multirow{3}{*}{OPD}
& Run 1
& \res{32.50}{0.85}
& \res{27.08}{1.01}
& \res{69.38}{0.78}
& \res{14.79}{0.73}
& \res{59.60}{0.64}
& \res{86.52}{0.36}
& $48.31$ \\

& Run 2
& \res{33.75}{1.08}
& \res{27.29}{0.77}
& \res{70.86}{0.95}
& \res{15.42}{0.63}
& \res{61.19}{0.86}
& \res{86.98}{0.26}
& $49.25$ \\

& Run 3
& \res{33.85}{0.78}
& \res{28.96}{0.72}
& \res{70.08}{0.95}
& \res{17.40}{0.59}
& \res{59.17}{0.25}
& \res{87.55}{0.27}
& $49.50$ \\

\rowcolor{summarygray}
& Mean
& \res{33.37}{0.75}
& \secondres{27.78}{1.03}
& \res{70.10}{0.74}
& \res{15.87}{1.36}
& \res{59.98}{1.07}
& \res{87.02}{0.51}
& \res{49.02}{0.63} \\

\midrule

% ============================================================
% ESR(4096)
% ============================================================

\multirow{3}{*}{ESR}
& Run 1
& \res{35.52}{0.83}
& \res{25.52}{0.87}
& \res{70.86}{1.11}
& \res{15.73}{0.78}
& \res{60.20}{0.88}
& \res{86.58}{0.29}
& $49.07$ \\

& Run 2
& \res{36.88}{0.88}
& \res{28.85}{0.99}
& \res{67.19}{0.75}
& \res{15.62}{0.59}
& \res{60.71}{0.29}
& \res{86.98}{0.24}
& $49.37$ \\

& Run 3
& \res{32.08}{0.98}
& \res{25.73}{1.01}
& \res{70.31}{1.00}
& \res{16.04}{0.64}
& \res{61.02}{0.29}
& \res{86.92}{0.16}
& $48.68$ \\

\rowcolor{summarygray}
& Mean
& \secondres{34.83}{2.47}
& \res{26.70}{1.87}
& \res{69.45}{1.98}
& \res{15.80}{0.22}
& \secondres{60.64}{0.41}
& \res{86.83}{0.22}
& \secondres{49.04}{0.35} \\

& \textit{$p$ vs.\ OPD}
& $0.210$
& $0.778$
& $0.679$
& $0.531$
& $0.207$
& $0.697$
& $0.482$ \\

\midrule

% ============================================================
% Prune-OPD
% Uses seeded runs s1/s2/s3; historical unseeded run excluded
% ============================================================

\multirow{3}{*}{Prune-OPD}
& Run 1
& \res{35.00}{1.15}
& \res{25.73}{0.95}
& \res{69.69}{0.90}
& \res{15.31}{0.72}
& \res{59.55}{0.73}
& \res{86.65}{0.38}
& $48.66$ \\

& Run 2
& \res{33.44}{1.14}
& \res{25.83}{1.02}
& \res{69.69}{0.98}
& \res{18.02}{0.70}
& \res{58.30}{0.40}
& \res{85.90}{0.26}
& $48.53$ \\

& Run 3
& \res{32.40}{1.07}
& \res{26.88}{0.81}
& \res{69.30}{0.94}
& \res{16.25}{0.70}
& \res{58.82}{0.38}
& \res{86.52}{0.29}
& $48.36$ \\

\rowcolor{summarygray}
& Mean
& \res{33.61}{1.31}
& \res{26.15}{0.64}
& \res{69.56}{0.23}
& \secondres{16.53}{1.38}
& \res{58.89}{0.63}
& \res{86.36}{0.40}
& \res{48.52}{0.15} \\

& \textit{$p$ vs.\ OPD}
& $0.402$
& $0.944$
& $0.747$
& $0.294$
& $0.892$
& $0.920$
& $0.852$ \\

\midrule

% ============================================================
% Relay-OPD
% ============================================================

\multirow{3}{*}{Relay-OPD}
& Run 1
& \res{31.46}{1.01}
& \res{27.40}{0.91}
& \res{71.17}{0.84}
& \res{16.46}{0.58}
& \res{60.07}{0.25}
& \res{87.25}{0.36}
& $48.97$ \\

& Run 2
& \res{31.04}{0.71}
& \res{26.04}{1.18}
& \res{71.02}{0.78}
& \res{15.31}{0.70}
& \res{59.68}{0.48}
& \res{86.92}{0.24}
& $48.34$ \\

& Run 3
& \res{31.15}{1.04}
& \res{28.44}{0.91}
& \res{69.14}{0.99}
& \res{14.27}{0.67}
& \res{58.69}{0.21}
& \res{87.47}{0.38}
& $48.19$ \\

\rowcolor{summarygray}
& Mean
& \res{31.22}{0.22}
& \res{27.29}{1.20}
& \secondres{70.44}{1.13}
& \res{15.35}{1.09}
& \res{59.48}{0.71}
& \secondres{87.21}{0.28}
& \res{48.50}{0.41} \\

& \textit{$p$ vs.\ OPD}
& $0.977$
& $0.687$
& $0.354$
& $0.684$
& $0.732$
& $0.307$
& $0.848$ \\

\midrule

% ============================================================
% SCOUT
% ============================================================

\multirow{3}{*}{\textbf{SCOUT}}
& Run 1
& \res{31.35}{1.15}
& \res{30.31}{0.93}
& \res{74.45}{0.85}
& \res{16.56}{0.69}
& \res{61.57}{0.64}
& \res{88.65}{0.30}
& $50.48$ \\

& Run 2
& \res{39.17}{0.99}
& \res{31.56}{0.92}
& \res{75.70}{0.80}
& \res{18.85}{0.73}
& \res{61.02}{0.11}
& \res{88.55}{0.43}
& $52.48$ \\

& Run 3
& \res{35.62}{0.72}
& \res{31.56}{0.70}
& \res{74.84}{0.97}
& \res{17.71}{0.53}
& \res{61.45}{0.54}
& \res{89.72}{0.27}
& $51.82$ \\

\rowcolor{scoutsummary}
& Mean
& \bestres{35.38}{3.92}
& \bestres{31.15}{0.72}
& \bestres{75.00}{0.64}
& \bestres{17.71}{1.15}
& \bestres{61.35}{0.29}
& \bestres{88.97}{0.65}
& \bestres{51.59}{1.02} \\

& \textit{$p$ vs.\ OPD}
& $0.235$
& \cellcolor[RGB]{255,230,180}$0.00652$
& \cellcolor[RGB]{255,230,180}$0.00125$
& \cellcolor[RGB]{255,230,180}$0.0752$
& \cellcolor[RGB]{255,230,180}$0.0720$
& \cellcolor[RGB]{255,230,180}$0.00828$
& \cellcolor[RGB]{255,230,180}$0.0140$ \\

\bottomrule
\end{tabular}
}
\end{table*}

%% file: Tables/table_skywork_7b_math_full.tex
\begin{table*}[t]
\centering
\caption{
Complete per-run results for the Skywork-OR1-Math-7B
$\rightarrow$ DeepSeek-R1-Distill-Qwen-1.5B setting on six math reasoning
benchmarks.
This table corresponds to Table~\ref{tab:deepseek-math-results}. \method~significantly outperforms OPD on 4 of 6 benchmarks and in average performance, which extends its generalizability beyond the Qwen model family.
}
\label{tab:math_skywork_to_1p7b_per_run}

\small
\setlength{\tabcolsep}{4.0pt}
\renewcommand{\arraystretch}{1.10}

\resizebox{\textwidth}{!}{
\begin{tabular}{llccccccc}
\toprule
\textbf{Method}
& \textbf{Run}
& \textbf{AIME24}
& \textbf{AIME25}
& \textbf{AMC23}
& \textbf{HMMT}
& \textbf{Olymp.}
& \textbf{MATH500}
& \textbf{Avg6} \\[-1pt]

&
\textit{Avg.@}
& \textit{32}
& \textit{32}
& \textit{32}
& \textit{32}
& \textit{4}
& \textit{8}
& -- \\

\midrule

\multirow{3}{*}{OPD}
& Run 1
& \res{37.92}{1.05}
& \res{30.52}{0.73}
& \res{77.27}{0.77}
& \res{19.17}{0.89}
& \res{60.46}{0.76}
& \res{89.00}{0.26}
& $52.39$ \\

& Run 2
& \res{38.33}{0.89}
& \res{29.90}{0.68}
& \res{78.44}{0.66}
& \res{19.06}{0.45}
& \res{60.80}{0.76}
& \res{89.15}{0.25}
& $52.61$ \\

& Run 3
& \res{37.08}{1.06}
& \res{29.58}{0.53}
& \res{78.05}{0.94}
& \res{18.96}{0.55}
& \res{60.41}{0.64}
& \res{89.10}{0.23}
& $52.20$ \\

\rowcolor{summarygray}
& Mean
& \secondres{37.78}{0.64}
& \bestres{30.00}{0.48}
& \secondres{77.92}{0.60}
& \bestres{19.06}{0.11}
& \secondres{60.56}{0.21}
& \secondres{89.08}{0.08}
& \secondres{52.40}{0.21} \\

\midrule

% ============================================================
% ESR(4096) / Early-stop OPD
% ============================================================

\multirow{3}{*}{ESR}
& Run 1
& \res{31.67}{1.20}
& \res{27.40}{0.83}
& \res{73.44}{0.82}
& \res{18.44}{0.67}
& \res{56.80}{0.61}
& \res{87.38}{0.36}
& $49.19$ \\

& Run 2
& \res{31.15}{1.02}
& \res{27.19}{0.75}
& \res{70.23}{0.96}
& \res{16.98}{0.53}
& \res{57.27}{0.69}
& \res{87.62}{0.32}
& $48.41$ \\

& Run 3
& \res{34.90}{1.15}
& \res{27.40}{0.79}
& \res{71.09}{0.88}
& \res{16.35}{0.59}
& \res{56.50}{0.51}
& \res{87.38}{0.40}
& $48.94$ \\

\rowcolor{summarygray}
& Mean
& \res{32.57}{2.03}
& \res{27.33}{0.12}
& \res{71.59}{1.66}
& \res{17.26}{1.07}
& \res{56.86}{0.39}
& \res{87.46}{0.14}
& \res{48.84}{0.40} \\

& \textit{$p$ vs.\ Frozen-OPD}
& $0.985$
& $0.994$
& $0.995$
& $0.960$
& $0.999$
& $0.998$
& $1.000$ \\

\midrule

% ============================================================
% Prune-OPD
% ONLY s1 / s2 / s3; historical unseeded run excluded.
% Aggregate recomputed from these three runs.
% ============================================================

\multirow{3}{*}{Prune-OPD}
& Run 1
& \res{22.60}{1.01}
& \res{19.17}{0.73}
& \res{56.88}{0.83}
& \res{10.94}{0.62}
& \res{39.93}{0.73}
& \res{72.12}{0.36}
& $36.94$ \\

& Run 2
& \res{20.52}{1.05}
& \res{18.44}{0.81}
& \res{56.48}{0.83}
& \res{11.46}{0.70}
& \res{41.22}{0.29}
& \res{72.28}{0.50}
& $36.73$ \\

& Run 3
& \res{21.67}{0.70}
& \res{18.85}{0.84}
& \res{56.95}{1.03}
& \res{11.04}{0.71}
& \res{40.32}{0.46}
& \res{71.47}{0.60}
& $36.72$ \\

\rowcolor{summarygray}
& Mean
& \res{21.60}{1.04}
& \res{18.82}{0.37}
& \res{56.77}{0.25}
& \res{11.15}{0.28}
& \res{40.49}{0.66}
& \res{71.96}{0.43}
& \res{36.80}{0.12} \\

& \textit{$p$ vs.\ Frozen-OPD}
& $1.000$
& $1.000$
& $1.000$
& $1.000$
& $1.000$
& $1.000$
& $1.000$ \\

\midrule

% ============================================================
% Relay-OPD
% ============================================================

\multirow{3}{*}{Relay-OPD}
& Run 1
& \res{32.81}{1.22}
& \res{28.65}{0.80}
& \res{74.84}{0.87}
& \res{16.04}{0.61}
& \res{58.30}{0.49}
& \res{88.47}{0.28}
& $49.85$ \\

& Run 2
& \res{35.10}{0.91}
& \res{27.19}{0.79}
& \res{72.97}{0.69}
& \res{17.29}{0.55}
& \res{58.35}{0.81}
& \res{88.35}{0.24}
& $49.88$ \\

& Run 3
& \res{36.25}{1.11}
& \res{29.17}{0.70}
& \res{76.09}{0.97}
& \res{15.73}{0.64}
& \res{60.03}{0.44}
& \res{88.08}{0.20}
& $50.89$ \\

\rowcolor{summarygray}
& Mean
& \res{34.72}{1.75}
& \secondres{28.34}{1.03}
& \res{74.63}{1.57}
& \res{16.35}{0.83}
& \res{58.89}{0.98}
& \res{88.30}{0.20}
& \res{50.21}{0.59} \\

& \textit{$p$ vs.\ Frozen-OPD}
& $0.963$
& $0.956$
& $0.976$
& $0.994$
& $0.958$
& $0.991$
& $0.992$ \\

\midrule

% ============================================================
% SCOUT
% ============================================================

\multirow{3}{*}{\textbf{SCOUT}}
& Run 1
& \res{40.42}{1.08}
& \res{29.90}{0.71}
& \res{82.03}{0.84}
& \res{18.33}{0.58}
& \res{61.62}{0.86}
& \res{90.08}{0.36}
& $53.73$ \\

& Run 2
& \res{39.90}{0.90}
& \res{30.52}{0.78}
& \res{80.23}{0.67}
& \res{18.96}{0.68}
& \res{62.13}{0.61}
& \res{90.05}{0.26}
& $53.63$ \\

& Run 3
& \res{39.06}{1.38}
& \res{29.58}{0.61}
& \res{81.17}{0.82}
& \res{18.75}{0.74}
& \res{61.75}{0.55}
& \res{90.00}{0.22}
& $53.39$ \\

\rowcolor{scoutsummary}
& Mean
& \bestres{39.79}{0.69}
& \bestres{30.00}{0.48}
& \bestres{81.14}{0.90}
& \secondres{18.68}{0.32}
& \bestres{61.83}{0.26}
& \bestres{90.04}{0.04}
& \bestres{53.58}{0.18} \\

& \textit{$p$ vs.\ Frozen-OPD}
& \cellcolor[RGB]{255,230,180}$0.0418$
& $0.500$
& \cellcolor[RGB]{255,230,180}$0.00504$
& $0.741$
& \cellcolor[RGB]{255,230,180}$0.0454$
& \cellcolor[RGB]{255,230,180}$0.00613$
& \cellcolor[RGB]{255,230,180}$<0.001$ \\

\bottomrule
\end{tabular}
}
\end{table*}

%% file: Tables/table_prefix0_math_full.tex
\begin{table*}[t]
\centering
\caption{
Per-run mathematical-reasoning results for the
\textbf{OPD + Teacher GRPO} control.
This control applies the same outcome-reward GRPO update to the teacher as
SCOUT, but the teacher rolls out from scratch without conditioning on a
student-generated prefix.
OPD and SCOUT are shown as three-run aggregates for reference.
For individual training runs, the smaller $\pm$ term denotes within-evaluation
standard error; in \textit{Mean} rows, it denotes standard deviation across
three training runs.
Avg. is the macro-average over the six displayed math benchmarks.
Best and second-best three-run aggregates are shown in
\textbf{bold} and \underline{underline}, respectively.
}
\label{tab:teacher_grpo_control_math_avg6}

\small
\setlength{\tabcolsep}{4.0pt}
\renewcommand{\arraystretch}{1.10}

\resizebox{\textwidth}{!}{
\begin{tabular}{llccccccc}
\toprule
\textbf{Method}
& \textbf{Run}
& \textbf{AIME24}
& \textbf{AIME25}
& \textbf{AMC23}
& \textbf{HMMT}
& \textbf{Olymp.}
& \textbf{MATH500}
& \textbf{Avg.} \\
& \textit{Avg.@}
& \textit{32}
& \textit{32}
& \textit{32}
& \textit{32}
& \textit{4}
& \textit{8}
& \textit{--} \\
\midrule

\multicolumn{9}{l}{
\textit{\textbf{(a) Qwen3-4B-Instruct-2507 $\rightarrow$ Qwen3-1.7B}}
} \\[2pt]

OPD
& Mean
& \secondres{33.82}{3.03}
& \res{24.31}{1.75}
& \secondres{73.93}{1.85}
& \secondres{16.04}{0.75}
& \secondres{60.15}{1.64}
& \res{87.03}{1.14}
& \secondres{49.21}{1.53} \\

\midrule

\multirow{3}{*}{OPD + Teacher GRPO}
& Run 1
& \res{29.17}{0.82}
& \res{23.65}{0.68}
& \res{73.05}{0.87}
& \res{13.96}{0.55}
& \res{58.56}{0.55}
& \res{86.67}{0.30}
& 47.51 \\

& Run 2
& \res{32.29}{1.05}
& \res{26.56}{0.73}
& \res{74.53}{0.96}
& \res{16.04}{0.66}
& \res{60.15}{0.72}
& \res{87.83}{0.17}
& 49.57 \\

& Run 3
& \res{32.50}{1.07}
& \res{27.40}{0.83}
& \res{71.95}{0.92}
& \res{17.71}{0.72}
& \res{59.42}{0.46}
& \res{87.12}{0.33}
& 49.35 \\

\rowcolor{summarygray}
& Mean
& \res{31.32}{1.86}
& \secondres{25.87}{1.97}
& \res{73.18}{1.29}
& \res{15.90}{1.88}
& \res{59.38}{0.80}
& \secondres{87.21}{0.58}
& \res{48.81}{1.13} \\

\midrule

\rowcolor{scoutsummary}
\textbf{SCOUT}
& Mean
& \bestres{35.76}{0.51}
& \bestres{28.82}{0.64}
& \bestres{75.76}{0.59}
& \bestres{17.22}{1.90}
& \bestres{62.38}{0.84}
& \bestres{88.41}{0.25}
& \bestres{51.39}{0.57} \\

\midrule
\multicolumn{9}{l}{
\textit{\textbf{(b) Qwen3-8B-DAPO $\rightarrow$ Qwen3-1.7B}}
} \\[2pt]

OPD
& Mean
& \res{33.37}{0.75}
& \res{27.78}{1.03}
& \res{70.10}{0.74}
& \res{15.87}{1.36}
& \res{59.98}{1.07}
& \res{87.02}{0.51}
& \res{49.02}{0.61} \\

\midrule

\multirow{3}{*}{OPD + Teacher GRPO}
& Run 1
& \res{38.54}{1.01}
& \res{32.40}{0.81}
& \res{74.45}{0.78}
& \res{17.60}{0.64}
& \res{61.10}{0.41}
& \res{88.90}{0.23}
& 52.17 \\

& Run 2
& \res{34.17}{0.99}
& \res{29.90}{0.71}
& \res{72.42}{0.81}
& \res{16.77}{0.69}
& \res{60.97}{0.66}
& \res{87.52}{0.24}
& 50.29 \\

& Run 3
& \res{32.29}{0.87}
& \res{31.77}{0.81}
& \res{71.56}{0.79}
& \res{16.35}{0.69}
& \res{59.85}{0.44}
& \res{87.67}{0.40}
& 49.92 \\

\rowcolor{summarygray}
& Mean
& \secondres{35.00}{3.21}
& \bestres{31.36}{1.30}
& \secondres{72.81}{1.48}
& \secondres{16.91}{0.64}
& \secondres{60.64}{0.69}
& \secondres{88.03}{0.76}
& \secondres{50.79}{1.21} \\

\midrule

\rowcolor{scoutsummary}
\textbf{SCOUT}
& Mean
& \bestres{35.38}{3.91}
& \secondres{31.15}{0.72}
& \bestres{75.00}{0.64}
& \bestres{17.71}{1.15}
& \bestres{61.35}{0.29}
& \bestres{88.97}{0.65}
& \bestres{51.59}{1.02} \\

\bottomrule
\end{tabular}
}
\end{table*}

%% file: Tables/table_prefix0_code_full.tex
% ============================================================
% Qwen3-4B-Instruct-2507 -> Qwen3-1.7B, Code
% Teacher-GRPO control ablation
% ============================================================

\begin{table*}[t]
\centering
\caption{
Per-run code-generation results for the
\textbf{OPD + Teacher GRPO} control in the
Qwen3-4B-Instruct-2507 $\rightarrow$ Qwen3-1.7B setting.
This control applies the same outcome-reward GRPO update to the teacher as
SCOUT, but the teacher rolls out from scratch without conditioning on a
student-generated prefix.
OPD and SCOUT are shown as three-run aggregates for reference.
For individual training runs, the smaller $\pm$ term denotes within-evaluation
standard error; in \textit{Mean} rows, it denotes standard deviation across
three training runs.
Best and second-best three-run aggregates are shown in
\textbf{bold} and \underline{underline}, respectively.
}
\label{tab:teacher_grpo_control_code_4b}

\small
\setlength{\tabcolsep}{7pt}
\renewcommand{\arraystretch}{1.10}

\begin{tabular}{llcccc}
\toprule
\textbf{Method}
& \textbf{Run}
& \textbf{LCB v5}
& \textbf{HumanEval+}
& \textbf{MBPP}
& \textbf{Avg.} \\
& \textit{Avg.@}
& \textit{4}
& \textit{16}
& \textit{8}
& \textit{--} \\
\midrule

OPD
& Mean
& \res{37.64}{0.05}
& \res{69.74}{0.68}
& \res{62.48}{0.29}
& \res{56.62}{0.28} \\

\midrule

\multirow{3}{*}{OPD + Teacher GRPO}
& Run 1
& \res{38.01}{0.30}
& \res{69.21}{0.50}
& \res{61.35}{0.49}
& 56.19 \\

& Run 2
& \res{39.40}{0.29}
& \res{72.18}{0.49}
& \res{65.80}{0.27}
& 59.13 \\

& Run 3
& \res{36.08}{0.24}
& \res{69.36}{0.60}
& \res{61.72}{0.63}
& 55.72 \\

\rowcolor{summarygray}
& Mean
& \secondres{37.83}{1.67}
& \secondres{70.25}{1.67}
& \secondres{62.96}{2.47}
& \secondres{57.01}{1.85} \\

\midrule

\rowcolor{scoutsummary}
\textbf{SCOUT}
& Mean
& \bestres{40.57}{2.51}
& \bestres{73.42}{0.88}
& \bestres{65.22}{2.62}
& \bestres{59.74}{1.88} \\

\bottomrule
\end{tabular}
\end{table*}

%% file: Tables/table_ablation_f_8b.tex
\begin{table*}[ht]
\centering
\caption{
Ablation on the teacher update interval $f$ with the Qwen3-8B-DAPO $\rightarrow$ Qwen3-1.7B student setting. Updating the teacher with $f\leq10$ consistently
improves over OPD, whereas the improvement becomes marginal or disappears
for $f\geq15$. Although $f=5$ achieves the highest average accuracy, $f=10$
retains most of the improvement with fewer teacher updates, providing a
favorable performance--computation trade-off.
}
\resizebox{\textwidth}{!}{%
\begin{tabular}{lccccccc}
\toprule
Model & AIME24 & AIME25 & AMC23 & HMMT & Olympiad
& MATH-500 & Mean \\
\midrule
Qwen3-1.7B
& $13.3{\scriptstyle\pm4.2}$
& $10.3{\scriptstyle\pm3.2}$
& $46.8{\scriptstyle\pm4.9}$
& $5.4{\scriptstyle\pm2.6}$
& $42.3{\scriptstyle\pm1.5}$
& $73.1{\scriptstyle\pm1.5}$
& $31.9$ \\
Qwen3-8B-DAPO
& $59.9{\scriptstyle\pm6.0}$
& $43.9{\scriptstyle\pm5.9}$
& $90.7{\scriptstyle\pm3.9}$
& $25.7{\scriptstyle\pm4.1}$
& $74.2{\scriptstyle\pm0.7}$
& $94.6{\scriptstyle\pm0.5}$
& $64.8$ \\
\midrule
OPD
& $33.5{\scriptstyle\pm6.5}$
& $28.0{\scriptstyle\pm5.5}$
& $69.2{\scriptstyle\pm4.9}$
& $16.4{\scriptstyle\pm4.1}$
& $59.9{\scriptstyle\pm0.4}$
& $87.2{\scriptstyle\pm0.9}$
& $49.0$ \\
\midrule
% NOTE: f=1 was run at lr 5e-7 and f=5 at lr 2.5e-6 -- resolve before submission.
SCOUT ($f{=}1$)
& $33.0{\scriptstyle\pm5.8}$
& $\underline{29.8}{\scriptstyle\pm4.6}$
& $\mathbf{76.1}{\scriptstyle\pm4.8}$
& $\underline{18.2}{\scriptstyle\pm3.9}$
& $\mathbf{62.9}{\scriptstyle\pm1.0}$
& $87.8{\scriptstyle\pm0.8}$
& $51.3$ \\
SCOUT ($f{=}5$)
& $\underline{38.4}{\scriptstyle\pm6.0}$
& $\mathbf{30.3}{\scriptstyle\pm5.4}$
& $\underline{75.8}{\scriptstyle\pm5.2}$
& $\mathbf{18.3}{\scriptstyle\pm4.1}$
& $\underline{62.7}{\scriptstyle\pm1.8}$
& $\mathbf{89.6}{\scriptstyle\pm0.6}$
& $\mathbf{52.5}$ \\
SCOUT ($f{=}10$)
& $\mathbf{39.2}{\scriptstyle\pm5.0}$
& $28.7{\scriptstyle\pm6.1}$
& $74.2{\scriptstyle\pm5.0}$
& $16.5{\scriptstyle\pm4.1}$
& $60.8{\scriptstyle\pm0.4}$
& $\underline{89.4}{\scriptstyle\pm0.7}$
& $\underline{51.5}$ \\
SCOUT ($f{=}15$)
& $34.3{\scriptstyle\pm4.9}$
& $27.7{\scriptstyle\pm5.0}$
& $70.7{\scriptstyle\pm5.8}$
& $15.3{\scriptstyle\pm4.2}$
& $60.0{\scriptstyle\pm1.0}$
& $87.2{\scriptstyle\pm1.1}$
& $49.2$ \\
SCOUT ($f{=}20$)
& $33.5{\scriptstyle\pm6.2}$
& $29.2{\scriptstyle\pm4.7}$
& $68.0{\scriptstyle\pm5.1}$
& $15.4{\scriptstyle\pm3.5}$
& $59.7{\scriptstyle\pm1.3}$
& $86.7{\scriptstyle\pm1.1}$
& $48.8$ \\
\bottomrule
\end{tabular}%
}
\label{tab:analysis-update-freq}
\end{table*}

%% file: Tables/table_optr_scout_math_full.tex
\begin{table*}[t]
\centering
\caption{
Full results for combining SCOUT with On-policy trust region OPD (OPTR) in the
Qwen3-4B-Instruct-2507 $\rightarrow$ Qwen3-1.7B math setting. OPTR performs better than OPD, and \method~further improves OPTR when combined with it. This shows that \method~is complementary to other loss-level OPD methods.
}
\label{tab:optr_scout_math_full}

\small
\setlength{\tabcolsep}{4.2pt}
\renewcommand{\arraystretch}{1.10}

\resizebox{\textwidth}{!}{
\begin{tabular}{llccccccc}
\toprule
\textbf{Method}
& \textbf{Run}
& \textbf{AIME24}
& \textbf{AIME25}
& \textbf{AMC23}
& \textbf{HMMT}
& \textbf{Olymp.}
& \textbf{MATH500}
& \textbf{Avg.} \\
& \textit{Avg.@}
& \textit{32}
& \textit{32}
& \textit{32}
& \textit{32}
& \textit{4}
& \textit{8}
& \textit{--} \\
\midrule

OPD
& Mean
& \res{33.82}{3.03}
& \res{24.31}{1.75}
& \res{73.93}{1.85}
& \res{16.04}{0.75}
& \res{60.15}{1.64}
& \res{87.03}{1.14}
& \res{49.21}{1.53} \\

\rowcolor{summarygray}
SCOUT
& Mean
& \secondres{35.76}{0.51}
& \bestres{28.82}{0.64}
& \secondres{75.76}{0.59}
& \bestres{17.22}{1.90}
& \secondres{62.38}{0.84}
& \secondres{88.41}{0.25}
& \secondres{51.39}{0.57} \\

\midrule

\multirow{3}{*}{OPTR}
& Run 1
& \res{33.02}{1.01}
& \res{24.06}{0.86}
& \res{76.56}{0.82}
& \res{17.50}{0.75}
& \res{61.14}{0.56}
& \res{86.92}{0.44}
& 49.87 \\

& Run 2
& \res{32.81}{0.95}
& \res{24.58}{0.74}
& \res{73.98}{0.85}
& \res{15.52}{0.66}
& \res{61.49}{0.43}
& \res{86.80}{0.49}
& 49.20 \\

& Run 3
& \res{34.90}{0.92}
& \res{26.25}{0.79}
& \res{75.08}{0.66}
& \res{16.67}{0.75}
& \res{61.66}{0.37}
& \res{87.35}{0.28}
& 50.32 \\

\rowcolor{summarygray}
& Mean
& \res{33.58}{1.15}
& \res{24.96}{1.14}
& \res{75.21}{1.29}
& \res{16.56}{0.99}
& \res{61.43}{0.27}
& \res{87.02}{0.29}
& \res{49.80}{0.56} \\

\midrule

\multirow{3}{*}{OPTR + SCOUT}
& Run 1
& \res{37.60}{0.98}
& \res{27.50}{0.86}
& \res{77.89}{0.78}
& \res{15.83}{0.67}
& \res{63.51}{0.98}
& \res{89.70}{0.40}
& 52.01 \\

& Run 2
& \res{37.08}{1.12}
& \res{27.19}{0.64}
& \res{80.00}{0.89}
& \res{18.65}{0.67}
& \res{63.64}{0.57}
& \res{89.48}{0.26}
& 52.67 \\

& Run 3
& \res{34.79}{0.73}
& \res{28.23}{0.86}
& \res{78.83}{0.77}
& \res{17.08}{0.70}
& \res{63.77}{0.22}
& \res{90.10}{0.22}
& 52.13 \\

\rowcolor{scoutsummary}
& Mean
& \bestres{36.49}{1.50}
& \secondres{27.64}{0.53}
& \bestres{78.91}{1.06}
& \secondres{17.19}{1.41}
& \bestres{63.64}{0.13}
& \bestres{89.76}{0.31}
& \bestres{52.27}{0.35} \\

\bottomrule
\end{tabular}
}
\end{table*}

%% file: Tables/table_optr_scout_code_full.tex
\begin{table*}[t]
\centering
\caption{
Full results for combining SCOUT with On-policy trust region OPD (OPTR) in the
Qwen3-4B-Instruct-2507 $\rightarrow$ Qwen3-1.7B code setting. When combined with \method, OPTR achieves further improvement, showing the same trend as Table~\ref{tab:optr_scout_math_full}. This proves that the complementary merit of \method~generalizes to the code domain.
}
\label{tab:optr_scout_code_full}

\small
\setlength{\tabcolsep}{7pt}

\begin{tabular}{llcccc}
\toprule
\textbf{Method} & \textbf{Run}
& \textbf{LCB v5}
& \textbf{HumanEval+}
& \textbf{MBPP}
& \textbf{Avg.} \\
& \textit{Avg.@}
& \textit{4}
& \textit{16}
& \textit{8}
& \textit{--} \\
\midrule

OPD
& Mean
& \res{37.64}{0.05}
& \res{69.74}{0.68}
& \res{62.48}{0.29}
& \res{56.62}{0.28} \\

\rowcolor{summarygray}
SCOUT
& Mean
& \res{40.57}{2.51}
& \secondres{73.42}{0.88}
& \secondres{65.22}{2.62}
& \secondres{59.74}{1.88} \\

\midrule

\multirow{3}{*}{OPTR}
& Run 1
& \res{41.42}{0.64}
& \res{73.67}{0.52}
& \res{63.77}{0.36}
& 59.62 \\

& Run 2
& \res{40.40}{0.26}
& \res{72.56}{0.94}
& \res{64.58}{0.56}
& 59.18 \\

& Run 3
& \res{41.42}{0.38}
& \res{72.79}{0.72}
& \res{63.45}{0.71}
& 59.22 \\

\rowcolor{summarygray}
& Mean
& \secondres{41.08}{0.59}
& \res{73.01}{0.58}
& \res{63.93}{0.58}
& \res{59.34}{0.24} \\

\midrule

\multirow{3}{*}{OPTR + SCOUT}
& Run 1
& \res{42.95}{0.46}
& \res{75.08}{0.49}
& \res{66.38}{0.50}
& 61.47 \\

& Run 2
& \res{40.99}{0.45}
& \res{73.48}{0.64}
& \res{63.58}{0.52}
& 59.35 \\

& Run 3
& \res{41.53}{0.55}
& \res{74.31}{0.74}
& \res{66.47}{0.18}
& 60.77 \\

\rowcolor{scoutsummary}
& Mean
& \bestres{41.83}{1.01}
& \bestres{74.29}{0.80}
& \bestres{65.47}{1.65}
& \bestres{60.53}{1.08} \\

\bottomrule
\end{tabular}
\end{table*}